\pdfoutput=1

\documentclass[11pt]{article}

\usepackage{acl}

\usepackage{physics}
\usepackage{comment}
\usepackage{rotating}
\usepackage[utf8]{inputenc}
\usepackage{times}
\usepackage{latexsym}
\usepackage{rotating,tabularx,siunitx}

\usepackage{float}
\usepackage{graphicx}
\usepackage{subfigure}
\usepackage{caption}
\usepackage{amsmath,amssymb}
\usepackage{mathtools}
\usepackage{booktabs}
\usepackage{multirow}
\usepackage{siunitx}
\usepackage{adjustbox}
\usepackage{pifont}
\usepackage[T1]{fontenc}

\usepackage[utf8]{inputenc}

\usepackage{microtype}

\newcommand{\arr}[1]{\textcolor{black}{#1}}

\title{Sentence Embedding Leaks More Information than You Expect: Generative Embedding Inversion Attack to Recover the Whole Sentence}

\author{{\bf Haoran Li}, {\bf Mingshi Xu}, {\bf Yangqiu Song}\\
Dept. of CSE, Hong Kong University of Science and Technology\\
 \texttt{\{hlibt, mxuax\}@connect.ust.hk},
 \texttt{yqsong@cse.ust.hk} \\
}

\begin{document}
\maketitle
\begin{abstract}
Sentence-level representations are beneficial for various natural language processing tasks.
It is commonly believed that vector representations can capture rich linguistic properties.
Currently, large language models (LMs) achieve state-of-the-art performance on sentence embedding.
However, some recent works suggest that vector representations from LMs can cause information leakage ~\cite{song-information-2020,Pan-2020-Privacy}.
In this work, we further investigate the information leakage issue and propose a generative embedding inversion attack (GEIA) that aims to reconstruct input sequences based only on their sentence embeddings.
Given the black-box access to a language model, we treat sentence embeddings as initial tokens' representations and \arr{train or fine-tune} a \arr{powerful} decoder model to decode the whole sequences directly.
We conduct extensive experiments to demonstrate that our generative inversion attack outperforms previous embedding inversion attacks in classification metrics and generates coherent and contextually similar sentences as the original inputs. 
\end{abstract}

\section{Introduction}
Sentence embeddings serve as ``universal embeddings'' that have been widely used for numerous natural language processing tasks,  e.g., text classification, question-answering, semantic retrieval, and other semantic similarity tasks~\cite{cer-etal-2017-semeval}.
Recently, embedding models exploit large pre-trained language models to achieve revolutionary performance~\cite{reimers-2019-sentence-bert,gao-2021-simcse}.
\arr{The notable improvement allows sentence embeddings to be directly used as inputs for downstream tasks.
}
However, when applying sentence embeddings in downstream tasks, unintended data disclosure may violate the legislation, result in fines and hinder individuals from contributing their data to service models.
For example, in some cases, such as legal document search, when submitting a query embedding to a service based on neural models, we may not want to leak the sensitive information in the query embedding.

\begin{figure*}[t]
\centering
\includegraphics[width=0.95\textwidth]{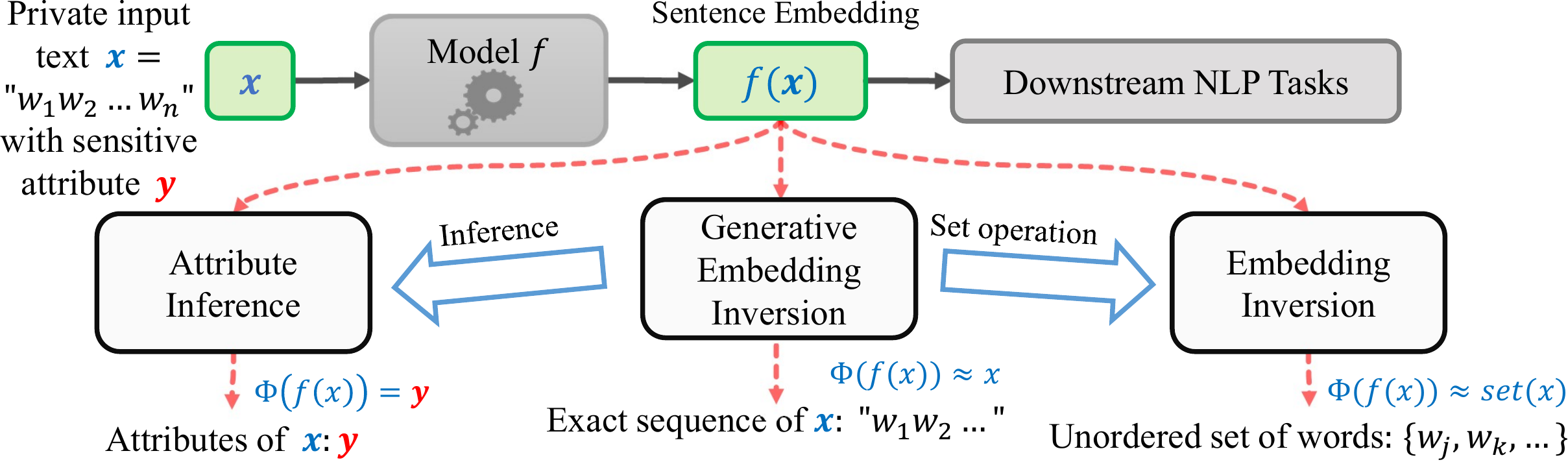}
\vspace{-0.1in}
\caption{
Overview of embedding inversion and attribute inference attacks on language models.
Both attacks can be conducted on the sentence embedding $f(x)$.
Previous embedding inversion attacks only predict sets of words while our generative embedding inversion attack is able to reconstruct actual input sequences.
}
\label{fig:intro_attacks}
\vspace{-0.2in}
\end{figure*}

Language models have already been proven to memorize training data and some private training data can be extracted~\cite{Carlini-Secret-2019,carlini-2021-extracting,thakkar-2021-understanding}.
Besides the memorization issue, representations learned from language models also inherently leak sensitive information and suffer from {\it attribute inference attacks}~\cite{Song2017MachineLM}.
Notably, for sentence embeddings, some of the words in the original sentence can be recovered, which is called {\it embedding inversion attacks}~\cite{song-information-2020,Pan-2020-Privacy}.
As shown in Figure~\ref{fig:intro_attacks}, both attribute inference attacks and embedding inversion attacks take sentence representations from language models as inputs.
For attribute inference attacks, the adversary builds a multi-layer perceptron (MLP) to infer the input sentences' private attributes (e.g., gender, race, and other identifiable personal information).
For embedding inversion attacks, existing approaches are viewed as  classification tasks that aim to recover partial keywords or unordered sets of original words from input sequences.

However, attribute inference attacks and existing embedding inversion attacks are not enough to explore the information leakage of sentence embeddings.
If the malicious adversary can recover original sentences from their embeddings, attribute inference attacks and previous embedding inversion attacks can be conducted subsequently.
Attribute inference attacks can be performed over the recovered sentences, which gives more flexibility for attack classifiers.
Moreover, existing embedding inversion attacks can be done by directly converting the recovered sentences to sets (bag-of-words).
Lastly, recovering the input sentences reveals more semantic information beyond the two attacks.

In this paper, we propose a generative embedding inversion attack (GEIA) to reconstruct input sentences given their embeddings from various language models.
Unlike previous embedding inversion attacks,
our attack is able to generate ordered sequences that share high contextual similarities with actual input sentences.
Our proposed attack advances preceding embedding inversion attacks from classification to generation.
Moreover, the generated sequences are mostly coherent and some are even verbatim text sequences from inputs.
Finally, our attack is adaptive and effective to various LM-based sentence embedding models regardless of the models' architectures and training methods.
We perform extensive experiments to demonstrate the effectiveness of our attacks on Sentence-BERT~\cite{reimers-2019-sentence-bert}, SimCSE-BERT/SimCSE-RoBERTa~\cite{gao-2021-simcse}, Sentence-T5~\cite{Ni-2021-SentenceT5} and MPNet~\cite{song-2020-mpnet}.
We also conduct experiments to show that our GEIA can even outperform previous attacks on classification metrics. 
Our contributions can be summarized as follows: \footnote{Code is publicly available at \url{https://github.com/HKUST-KnowComp/GEIA}.}

(1) To the best of our knowledge, we are the first to treat the embedding inversion attack as a generation task rather than a classification task.
This allows our attack to reconstruct ordered sequences.

(2)
Our GEIA can be adaptive to various embedding models with different model architectures from BERT to T5
and training algorithms like contrastive learning and siamese networks.

(3) \arr{We conduct extensive experiments to show the effectiveness of GEIA}.
Our results suggest that current state-of-the-art embedding models are vulnerable to GEIA.
\section{Related Works}
\textbf{Sentence embedding with language models.}
Sentence embeddings aim to train universal vector representations that can handle downstream tasks.
Earlier works learn sentence representations by exploiting encoder-decoder architectures to predict surrounding sentences ~\cite{Kiros-skip-2015,Gan-Unsupervised-2017} and autoencoder models to reconstruct original sentences ~\cite{hill-etal-2016-learning,zhang-etal-2018-learning-universal}.
Recent works turn to deeper and more complex transformer-based neural architectures like BERT ~\cite{devlin-etal-2019-bert} and RoBERTa ~\cite{liu-2019-roberta} to further improve sentence representations.
Sentence-BERT ~\cite{reimers-2019-sentence-bert} proposed a siamese network to improve the efficiency and performance of BERT representations.
SimCSE ~\cite{gao-2021-simcse} applied contrastive learning on BERT by self-predicting with dropout.
Currently, Sentence-T5 ~\cite{Ni-2021-SentenceT5} exploits T5 ~\cite{2020t5} and contrastive learning to further improve embeddings on various tasks of semantic textual similarity (STS) ~\cite{conneau-kiela-2018-senteval}.

\begin{figure*}[t]
\centering
\includegraphics[width=1\textwidth]{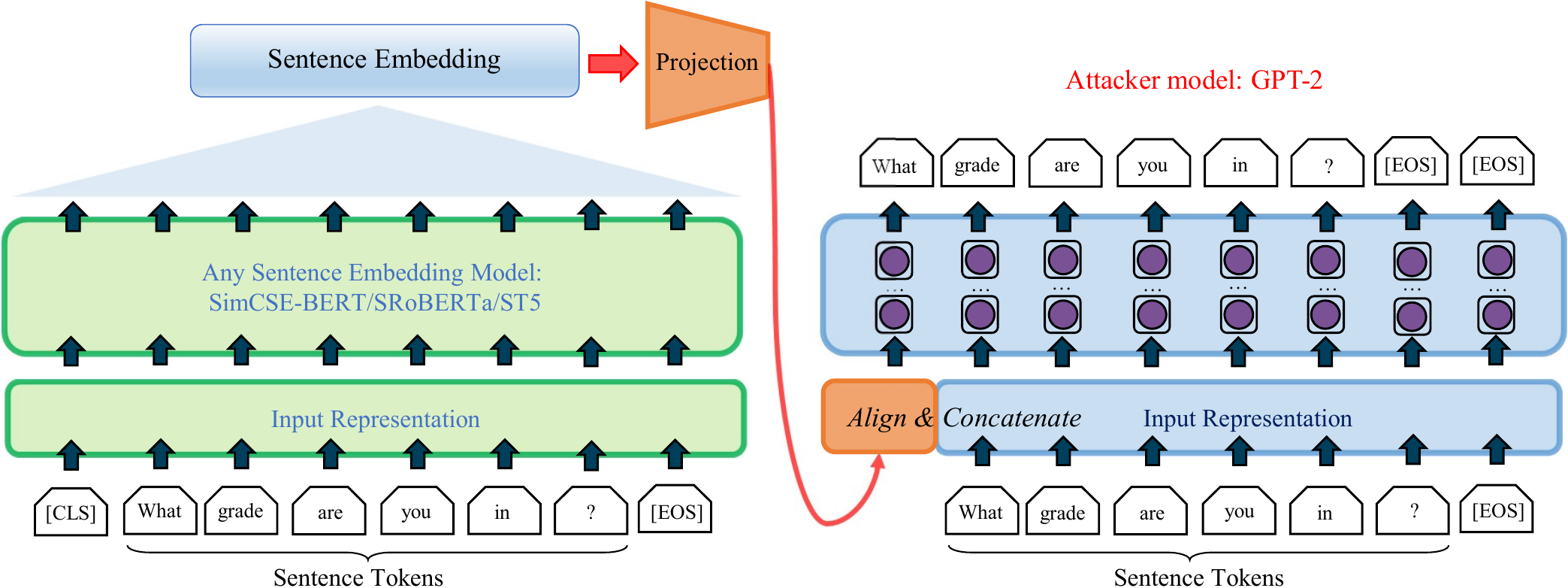}
\vspace{-0.2in}
\caption{
Model architecture for GEIA.
The sentence embedding can be embedded from arbitrary pretrained sentence embedding models.
The sentence embeddings are projected to the exact dimension of input token representations.
After projection, the projected embeddings are concatenated with input representations to train the attacker.
During inference, the sentence embeddings are fed as the initial token representations to decode corresponding inputs.
}
\label{fig:model}
\vspace{-0.2in}
\end{figure*}

\textbf{Privacy leakage on language models.}
Even though language models bring dramatic improvements to sentence representations, there are rising privacy concerns on language models.
~\citet{carlini-2021-extracting} showed that language models tended to memorize training data and performed training data extraction attacks to recover private training data.
~\citet{Gupta-2022-Recovering} studied deep gradient leakage \cite{Zhu-2019-Deep,Zhao-2020-iDLG} in language models and extracted training texts from aggregated gradients.
Besides the memorization issue, sentence embeddings from language models also encode private information that can easily be inferred by the adversary.
\citet{Pan-2020-Privacy} recovered partial fixed patterns and keywords from language models' sentence representations by querying language models with external annotated datasets.
Similarly, \citet{song-information-2020} performed attribute inference attacks and embedding inversion attacks to predict unordered sets of words from sentence embeddings.
Though their attack setups are similar to our work, they simply view embedding inversion attacks as classification problems and cannot reconstruct input sequences.
In this work, we reformulate the embedding inversion attack as a generation task and aim to invert ordered and informative sentences.
Moreover, our proposed attack outperforms previous attacks even on classification metrics.
\section{Embedding Inversion Attacks on LMs}

\subsection{Motivation}
\label{sec:Motivation}

\arr{To show how embedding inversion attacks compromise privacy, we hereby define the breach of privacy first.
Unintended or unauthorized data disclosure is regarded as privacy leakage during intended uses of the system.
In our scenario, the intended uses refer to obtaining sentence embeddings from target embedding models while recovering original sentences through external attacker models (other than intended decoders of autoencoders) is unintended and unauthorized.
When we use sentence embeddings for downstream applications and do not want to leak the original sentences, for example, personalized search in a neuralized document search engine in sensitive domains such as legal, medical, and financial domains, sensitive sentences may be unauthorizedly disclosed by curious service providers. 
Such unintended data disclosure may violate the legislation, incur fines and depress the service users.
}

\subsection{Problem Formulation}
\label{sec:Problem Formulation}
Given a sensitive input text sequence $x$ and a pre-trained language model $f$ on sentence representation, embedding inversion attacks aim to reconstruct the input $x$ from its sentence embedding $f(x)$.
More specifically,  the victim embedding model $f$ is already pre-trained and its parameters are frozen.
The adversary cannot update or modify the victim model's architecture and parameters.
Instead, the adversary holds an auxiliary dataset $D_{aux}$ that has a similar distribution to the data during attacks and attempts to build an external attacker model $\Phi$ to learn the inverse mapping $f^{-1}$ such that:
\begin{equation}
\label{eqn:attacker goal}
\small
\Phi (f(x)) \approx f^{-1} (f(x)) = x.
\end{equation}
Due to the fact that sentence embeddings are commonly aggregated by pooling operations on final hidden representations of individual tokens, the mapping $f$ that maps $x$ to $f(x)$ is inherently not one-to-one (injective).
\arr{Thus, it is impossible for attacker model $\Phi$ to behave like the inverse mapping $f^{-1}$.
And it remains challenging for $\Phi$ to approximate $f^{-1}$.}

\subsection{Existing Embedding Inversion Attacks}
\label{sec:Embedding Inversion}
Previously, \citet{song-information-2020} considered both white-box and black-box attacks to recover sets of words from short input text sequences.
For the white-box embedding inversion attack, it is assumed that the adversary can access the embedding model $f$'s parameters and architecture.
To make full use of the free access to $f$'s parameters, the adversary first builds a model $M$ that maps the deep layer representation $f(x)$ to its shallow layer representation.
Then the recovered set of words $\hat{x}$ are inferred from the shallow layer representation via continuous
relaxation \cite{Jang-Categorical-2017}.
For the black-box embedding inversion attack, the adversary can only query embedding model $f$ with $x$ from auxiliary data $D_{aux}$ to obtain its embedding $f(x)$.
Then the adversary directly learns $\Phi$ from $(f(x),x)$ pairs through multi-label classification with MLP or multi-set prediction with RNN.

\subsection{Limitations of Previous Approaches}
\label{sec:Limitations}
Though these aforementioned embedding inversion attacks may recover some words from corresponding embeddings, existing approaches have several limitations.
First, in the later experiment section, we show that existing embedding inversion attacks (multi-label classification and multi-set prediction) mainly predict insensitive stop words.
Such attacks are incapable of inverting informative contents from sentences' embeddings and therefore existing attacks are ineffective.
Second, predicting sets of words cannot handle word repetitions well in a text sequence.
Taking the epistrophe from the Bible as one example: ``When I was a child, I spoke as a child, I understood as a child...''
Simply predicting the token ``child'' can never capture the affluent linguistic properties.
Lastly, such predicted sets are also orderless and semantic information of ordering is permanently lost.
Taking the sentence ``Alice likes Bob'' as one example, even though we obtain the exact set of words: \{Alice, likes, Bob\}, we may still get the wrong meaning ``Bob likes Alice.''
As a result, existing approaches for recovering a set of words in a given sentence embedding are not as vicious as they claim.

\subsection{Generative Embedding Inversion Attacks}
\label{sec:Improved Embedding Inversion}
To overcome the above limitations, we propose GEIA which attempts to generate sentences that are contextually similar to actual inputs. 
We follow the black-box setup that the adversary can only query the victim language model whose architecture and parameters are inaccessible.
Intuitively, high-quality sentence embeddings encode rich linguistic properties about these sentences and a powerful generative decoder may utilize sentence embeddings to reconstruct original sentences.
To invert a sequence given its embedding, we propose using a generative attacker model to decode words based on the embedding and previous contexts.
As illustrated in Figure~\ref{fig:model}, the attacker model $\Phi$ can exploit powerful language models like GPT-2 \cite{radford-2019-language} to generate a sequence word by word from any given sentence embedding.

To train the attacker model, language modeling is applied with teacher forcing \cite{Williams-1989-teacher} to generate a text sequence word by word:
\begin{equation}
\label{eqn:LM}
\small
L_{\Phi} (x;\theta_{\Phi}) =
-\sum_{i=1}^{u} \log(\text{Pr}(w_i|f(x),w_0,w_1,...,w_{i-1})),
\end{equation}
where $x=$``$w_0 w_1 ... w_{u-1}$'' is a sentence of length $u$ from the auxiliary data $D_{aux}$ and $f(x)$ is the sentence embedding of $x$.

\arr{Unlike conventional encoder-decoder embedding models that intentionally and jointly train a decoder to strengthen the encoder, our GEIA solely trains the decoder based on the pre-trained and frozen embedding model $f$.
By contrast, we treat the sentence embedding $f(x)$ as the initial token representation before feeding the first word token $w_0$ of $x$.}
Here, the token representation means the input to the first transformer block.
If there is a size mismatch between the sentence embedding $f(x)$ and the attacker model's token representation, we apply one fully connected layer to align $f(x)$ to be the same size as the attacker model's token representation.
\arr{We use $Align(f(x))$ to denote the aligned sentence embedding and $\Phi_{emb}(w_i)$ to denote the representation of token $w_i$.
We concatenate $Align(f(x))$ to the left of the tokens' representation to obtain the attacker's input of $x$: 
\text{
\footnotesize{
$[Align(f(x)), \Phi_{emb}(w_0),\Phi_{emb}(w_1),...,\Phi_{emb}(w_{u-1})]$}
}.
This input bypasses $\Phi$'s embedding layer and is directly fed to $\Phi$'s first transformer block for text generation.
As illustrated in Equation~\ref{eqn:LM}, the attacker $\Phi$ manages to maximize the probability of the target sequence [$w_0,w_1,...,w_{u-1},$<eos>] given the input by minimizing the cross-entropy loss at each time step, where the <eos> indicates the special end of sentence token.
}

For inference, the attacker $\Phi$ decodes the first token from the aligned sentence embedding.
Then tokens are generated iteratively from  previous contexts with the sentence embedding till <eos> is reached.
We use $\Phi (f(x))$ to denote the whole generated sequence. 

\section{Experiments}

\begin{table*}[!htbp]
\centering
\small
  \begin{tabular}{c l  cccc | ccc | ccc}
    \toprule
    \multirow{2}{*}{Data} &
    \multirow{2}{*}{Victim Model} &
      \multicolumn{4}{c|}{MLC} &
      \multicolumn{3}{c|}{MSP} &
      \multicolumn{3}{c}{GEIA} \\
      {} &{} & {Threshold} & {Pre} & {Rec} & {F1} & {Pre} & {Rec} & {F1} & {Pre} & {Rec} & {F1}  \\
      \midrule
      \multirow{5}{*}{PC} & 
      SRoBERTa        & 0.20 & 33.42 & 26.79 & 29.74  & 43.39 & 38.12 & 40.59 & \arr{\textbf{58.41}} & \arr{\textbf{48.91}} & \arr{\textbf{53.24}} \\
    & SimCSE-BERT     & 0.50 & 24.77 & 21.36 & 22.94  & 42.23 & 37.10 & 39.50 & \arr{\textbf{66.95}} & \arr{\textbf{59.69}} & \arr{\textbf{63.11}} \\
    & SimCSE-RoBERTa  & 0.50 & 54.58 & 28.15 & 37.14  & 38.79 & 34.08 & 36.29 & \arr{\textbf{64.27}} & \arr{\textbf{56.66}} & \arr{\textbf{60.22}} \\
    & ST5             & 0.10 & 22.93 & 38.17 & 28.65  & 41.69 & 36.63 & 38.99 & \arr{\textbf{67.46}} & \arr{\textbf{58.26}} & \arr{\textbf{62.53}} \\
    & MPNet           & 0.20 & 33.91 & 27.39 & 30.30  & 39.23 & 34.46 & 36.69 & \arr{\textbf{62.64}} & \arr{\textbf{53.51}} & \arr{\textbf{57.72}} \\
      \midrule
      \midrule
      \multirow{5}{*}{QNLI} & 
    SRoBERTa            & 0.20  & 44.73 & 19.68 & 27.33    & \textbf{47.42} & 22.47 & 30.49 & \arr{43.81} & \arr{\textbf{27.19}} & \arr{\textbf{33.56}} \\
    & SimCSE-BERT       & 0.60  & 10.48 & 3.90  & 5.69   & 46.43 & 22.00 & 29.85 & \arr{\textbf{48.78}} & \arr{\textbf{29.49}} & \arr{\textbf{36.76}} \\
    & SimCSE-RoBERTa    & 0.75 & 28.74 & 10.10 & 14.95  & \textbf{52.57} & 24.90 & 33.80 & \arr{48.62} & \arr{\textbf{29.26}} & \arr{\textbf{36.53}} \\
    & ST5               & 0.20  & 42.26 & 19.83 & 27.00  & \textbf{48.50} & 22.98 & 31.18 & \arr{47.42} & \arr{\textbf{28.43}} & \arr{\textbf{35.55}} \\
    & MPNet             & 0.45 & \textbf{53.25} & 10.29 & 17.24  & 47.18 & 22.35 & 30.33 & \arr{44.89} & \arr{\textbf{27.74}} & \arr{\textbf{34.29}} \\
    \bottomrule
  \end{tabular}
  \vspace{-0.05in}
  \caption{
  \label{tab:qnli classification}
Embedding inversion performance comparison of multi-label classification, multi-set prediction and generative embedding inversion on classification metrics. 
The evaluations are done on the PersonaChat and QNLI datasets.
The token-level micro-averaged precision, recall and F1 are reported.
Precision (Pre), recall (Rec) and F1 are measured in \%.
High Pre, Rec and F1 indicate good attacking performance on classification.
}
\vspace{-0.1in}
\end{table*}

\subsection{Experimental Settings}
\label{exp:Experimental Settings}
\textbf{Datasets.}
Most sentence embedding models are trained on question-answer pairs (semi-supervised tasks) and natural language inference (supervised tasks) datasets.
For our experiments, we evaluate the attacking performance on 2 datasets.
The first dataset is PersonaChat (PC) dataset ~\cite{zhang-etal-2018-personalizing} that collects the open-domain chit-chat between two speakers given assigned personas.
Most personas are reflected in corresponding utterances, and some of them can be sensitive and private. 
The second dataset is QNLI ~\cite{wang-etal-2018-glue} converted from Stanford Question Answering Dataset ~\cite{rajpurkar-etal-2016-squad}.
The QNLI dataset is collected from Wikipedia articles that include domain knowledge.
Such domain knowledge can be challenging for inversion attacks.
For evaluation, we use their training data as the auxiliary dataset to train the attacker model and report their testing performance, respectively.
A summary of two datasets is shown in Table~\ref{tab:dataset-table}.

\begin{table}
\centering
\resizebox{0.48 \textwidth}{!}{
\begin{tabular}{lcc}
\hline
\textbf{Stat Type} & \textbf{PersonaChat} & \textbf{QNLI}\\
\hline
Task        & Dialog        & NLI \\
Domain      & Chit-chat     & Wikipedia \\
Sentences   & 162,064     & 220,412 \\
Train/dev/test split ratio & 82:9:9 & 95:0:5\\
Unique named entities & 1,425     & 46,567  \\
Avg. sentence length & 11.71     & 18.25 \\

\hline
\end{tabular}
}
\vspace{-0.05in}
\caption{\label{tab:dataset-table}
Statistics of datasets.
}
\vspace{-0.25in}
\end{table}

\textbf{Sentence Embedding Models.}
To perform embedding inversion attacks, we consider the following 5 victim sentence embedding models: \arr{Sentence-BERT~\cite{reimers-2019-sentence-bert}, SimCSE-BERT/SimCSE-RoBERTa~\cite{gao-2021-simcse}, Sentence-T5~\cite{Ni-2021-SentenceT5} and MPNet~\cite{song-2020-mpnet}.} All the embedding models' parameters are frozen and the pre-trained weights in their original GitHub repository are used.
\arr{Details and checkpoints of victim models are reported in Appendix~\ref{sec:Details of Victim}.}

\textbf{Attacker Models of Embedding Inversion.}
To compare with previous embedding inversion attacks, we implement two baseline attacker models proposed by ~\citet{song-information-2020}.

$\bullet$ Multi-label classification (MLC). 
Given the embedding $f(x)$ of sentence $x$. The adversary uses a simple MLP with binary cross entropy loss to predict words of $x$ over the whole vocabulary.

$\bullet$ Multi-set prediction (MSP). 
MLC independently predicts words of $x$ and ignores the dependency between words of a sequence. 
Multi-set prediction utilizes RNN architecture with multi-set prediction loss. 
We use the same multi-set objective that maximizes the probability of the set of tokens not predicted at the current time step as the previous work ~\cite{Welleck-2018-Multiset}.

$\bullet$ GEIA. As shown in Figure~\ref{fig:model}, our inversion attack can be regarded as sequence generation instead of set prediction.
\arr{We train a GPT-2 as the attacker model from random weights.
The random initialization makes a fair comparison between GEIA and previous baselines since both baselines also train from scratch. 
During our experiments, beam search decoding is applied for sentence recovery of GEIA.
We also experiment on several attacker models with pre-trained weights and decoding methods.
We put all the detailed evaluation results in the Appendix.
}

\textbf{Evaluation Metrics.}
To make a fair comparison, our evaluation considers both classification and generation metrics.

The classification metrics include token-level micro precision/recall/F1 for previous embedding inversion and our generative inversion attacks.
In addition, to study whether the recovered tokens are informative or not, we treat named entities as sensitive information and stop words as non-informative tokens.
We use the named entity recovery ratio (NERR) 
to measure the percentage of name entities that can be retrieved from input texts.
We also propose the stop word ratio (SWR) to calculate the percentage of stop words for given sentences.
Vicious embedding inversion attacks can achieve high NERRs with similar SWRs of original sentences.


Different from previous works, our GEIA can generate sequences instead sets of words.
So we also evaluate GEIA on the sentence-level generation metrics.
We apply ROUGE ~\cite{lin-2004-rouge}, BLEU ~\cite{Papineni-2002-BLEU}, and embedding similarity (ES) to measure the similarity between input sentences and inverted sequences.
ROUGE and BLEU both measure similarity based on n-grams.
ROUGE focuses on recall: how much the n-grams in the inputs are recovered.
BLEU measures precision: how much the n-grams inverted are correct.
The embedding similarity exploits ``sentence-t5-xxl'' to compute cosine similarity for the semantic similarity.
In addition, we use perplexity (PPL) of fine-tuned ``gpt2-large'' to measure the fluency of generated sentences.

\begin{table*}[t]
\centering
\small
  \begin{tabular}{c l  cccc | ccc}
    \toprule
    \multirow{2}{*}{Data} &
    \multirow{2}{*}{Victim Model} &
      \multicolumn{4}{c|}{SWR} &
      \multicolumn{3}{c}{NERR}  \\
     {}  & {} & {Test Set} & {MLC} & {MSP} & {GEIA} & {MLC} & {MSP} & {GEIA}   \\
      \midrule
      \multirow{5}{*}{PC} 
    & SRoBERTa        &               & +38.80 & +25.69 & \arr{\textbf{-05.01}} & 00.05 & 00.05 & \arr{\textbf{27.62}} \\
    & SimCSE-BERT     &               & -20.50 & +27.58 & \arr{\textbf{-06.10}} & 00.03 & 00.08 & \arr{\textbf{55.57}} \\
    & SimCSE-RoBERTa  & 61.06         & \textbf{+00.52} & +34.49 & \arr{-06.14} & 00.87 & 00.15 & \arr{\textbf{52.56}} \\
    & ST5             &               & +33.66 & +30.99 & \arr{\textbf{-05.70}} & 00.05 & 00.05 & \arr{\textbf{44.66}} \\
    & MPNet           &               & +38.83 & +30.54 & \arr{\textbf{-05.31}} & 00.05 & 00.05 & \arr{\textbf{32.50}} \\
    \midrule
    \midrule
      \multirow{5}{*}{QNLI} 
    & SRoBERTa        &               & +56.83 & +40.55 & \arr{\textbf{+05.14}}  & 01.06 & 02.12 & \arr{\textbf{15.12}} \\
    & SimCSE-BERT     &               & -18.79 & +40.97 & \arr{\textbf{+04.04}}  & 00.10 & 01.84 & \arr{\textbf{16.53}} \\
    & SimCSE-RoBERTa  & 38.13         & \textbf{-00.06} & +37.39 & \arr{+03.65}  & 00.82 & 02.50 & \arr{\textbf{18.16}} \\
    & ST5             &               & +56.77 & +39.35 & \arr{\textbf{+04.45}}  & 01.06 & 02.09 & \arr{\textbf{14.98}} \\
    & MPNet           &               & +61.87 & +41.16 & \arr{\textbf{+04.31}}  & 00.70 & 01.97 & \arr{\textbf{15.03}} \\
    \bottomrule
  \end{tabular}
  \vspace{-0.05in}
  \caption{\label{tab:nerr}
Embedding inversion performance on stop word rate (SWR) and named entity recovery ratio (NERR). 
NERR and SWR are measured in \%.
For SWR, we report the SWRs of testing data as baselines and the differences between baselines' SWRs and SWRs of various embedding inversion attacks.
A high NERR with a relatively low SWR difference suggests that the recovered tokens are informative, while a high SWR with a low NERR indicates that the attack is unsuccessful despite good classification performance.
}
\vspace{-0.1in}
\end{table*}
\begin{figure}[t]
\centering
\includegraphics[width=1\linewidth]{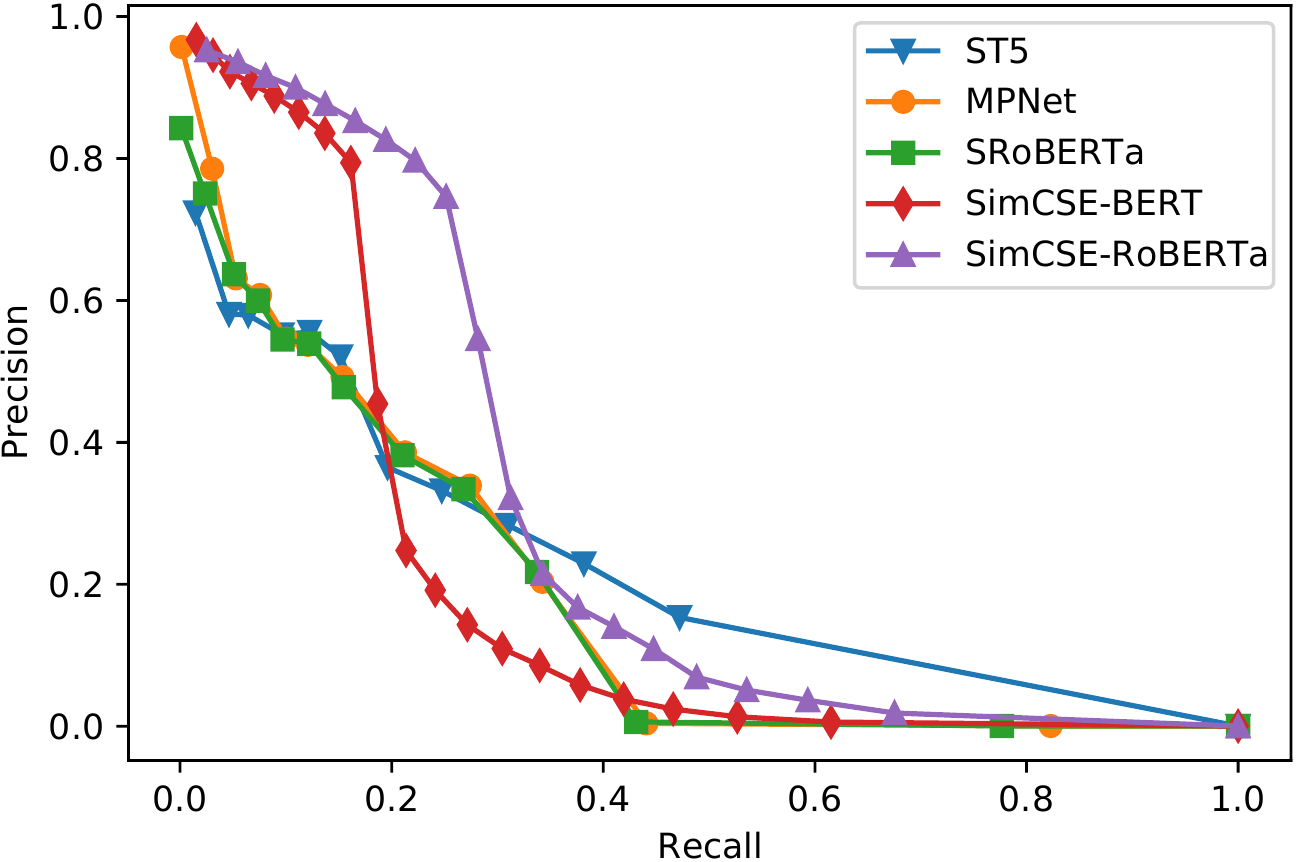}
\caption{
Precision-recall curve of MLC on the PersonaChat dataset.
}
\label{fig:prcurve}
\vspace{-0.15in}
\end{figure}

\subsection{Evaluation on Classification}
\label{exp:Performance}
Firstly, we compare the attacking performance between previous inversion baselines (MLC and MSP) and our GEIA on classification metrics.

Table~\ref{tab:qnli classification} shows the attacking performance of previous inversion attacks and our GEIA on the testing data.
We report token-level micro precision (Pre), recall (Rec), and F1 for all 3 attacks.
For multi-label classification, we carefully tune the binary thresholds based on the validation set and report the thresholds with the highest F1 scores.
By comparison, it can be seen that our \arr{GEIA} outperforms MLC and MSP.
Except MSP has a slightly higher F1 on SimCSE-RoBERTa of the QNLI dataset, our GEIA has the highest recall and F1 on 5 victim embedding models for both datasets.
On the PersonaChat dataset, our GEIA has the dominating performance with F1 scores of around \arr{0.58} while MSP has F1 scores of around 0.4 and MLC has an average F1 score of no more than 0.3.
These results show that our generative inversion attack has better attacking performance on classification tasks even though it is designed for generating sequences.

Despite tenable classification performance, previous MLC and MSP have several limitations.
For MLC, we find that most predicted tokens' probabilities are close to the reported thresholds and it is hard to distinguish tokens on MLC.
For MSP, though it performs better than MLC, it cannot handle long time steps well.
And the high precision on the QNLI dataset also benefits from small time steps.
In addition, they both tend to invert uninformative stop words and most inverted results include no sensitive content.
The subsequent experiments can help verify these stated limitations.


\subsection{Precision-Recall Trade-off of MLC}
\label{exp:Evaluation on Previous}
To better understand limitations of the baseline classifier's performance on embedding inversion, we draw the precision-recall curve for MLC with different thresholds.
Figure~\ref{fig:prcurve} shows the attack results for all victim embedding models where the individual result is marked by the marker.
Most data points are clustered in the upper-left zone of the figure.
The result indicates that MLC frequently leads to high precision with low recall.
As the threshold increases, for attacking performance of all five models, the precision drops dramatically compared with a minor increase in the recall. 

Regardless of the high precision, most areas under the curves are small.
The small areas imply that MLC is not a good classification model.
High precision means that MLC makes most predictions correctly while low recall means that MLC returns very few predictions.
Since the distribution among tokens is highly imbalanced and NNs may easily overfit the training data,
it remains unknown whether the predicted tokens are sensitive or not.

\begin{table*}[!htbp]
\centering
\small
  \begin{tabular}{l  cccccc  ccc}
    \toprule
    \multirow{2}{*}{} &
    \multicolumn{1}{c}{\multirow{2}{*}{PPL}} &
    \multicolumn{1}{c}{\multirow{2}{*}{ES}} &
      \multicolumn{2}{c}{ROUGE} &
      \multicolumn{3}{c}{BLEU} &
       \\
      & {} & {} & {ROUGE-1} & {ROUGE-L} & {BLEU-1} & {BLEU-2} & {BLEU-4} & \\
      \midrule
    SRoBERTa        & \arr{4.99} & \arr{88.07} & \arr{59.54} & \arr{56.04} & \arr{35.47} & \arr{20.37} & \arr{15.66} \\
    SimCSE-BERT     & \arr{6.29} & \arr{91.28} & \arr{72.38} & \arr{65.33} & \arr{46.93} & \arr{28.99} & \arr{22.85} \\
    SimCSE-RoBERTa  & \arr{5.98} & \arr{91.33} & \arr{68.78} & \arr{62.42} & \arr{43.41} & \arr{25.66} & \arr{19.82} \\
    ST5             & \arr{5.90} & \arr{91.47} & \arr{70.72} & \arr{65.45} & \arr{44.52} & \arr{27.83} & \arr{21.99} \\
    MPNet           & \arr{5.64} & \arr{89.27} & \arr{65.08} & \arr{60.39} & \arr{40.04} & \arr{23.83} & \arr{18.54} \\
    \midrule
    GPT-2 (w/o context)& 6.32 & 63.24 & 13.16 & 12.93 & 9.86 & 0.29 & 0.15 \\
    GPT-2 (w/ context) & 9.62  & 68.85 & 22.86 & 22.02 & 19.82 & 4.99 & 2.78 \\
    \bottomrule
  \end{tabular}
  \vspace{-0.05in}
  \caption{\label{tab:generation}
Evaluation on generation quality of generative embedding inversion attacks on victim embedding models and baseline models.
ES refers to embedding similarity and PPL refers to the perplexity of a fine-tuned GPT-2 model.
Embedding Similarity, ROUGE, and BLEU are measured in \%.
The two ``GPT-2'' models serve as baselines to generate the sequence given the first input token with/without context.
}
\vspace{-0.15in}
\end{table*}
\subsection{What Types of Tokens are Inverted?}
\label{exp:NERR and SWR}
Though the classification performance is evaluated, the informativeness of inverted tokens remains unknown.
Here we study the sensitivity of recovered tokens based on the named entity recovery ratio (NERR) and stop word rate (SWR).
A menacing embedding inversion attack can recover most named entities of original sequences.
If the recovered tokens are mostly stop words, the embedding inversion should be regarded as a failure.

Table~\ref{tab:nerr} includes NERRs and SWRs for MLC, MSP and GEIA.
For NERR, both MLC and MSP can only recover less than 3\% named entities from sentence embeddings for all situations.
\arr{On average,} our GEIA can invert around \arr{40\%} named entities on the PersonaChat dataset and around \arr{15\%} named entities on the QNLI dataset.
The results on NERRs suggest that our GEIA can indeed recover sensitive content while previous baselines fail to capture informative entities from embedding inversion attacks.
For SWR, we first report the SWRs of the original datasets and then calculate the SWR differences between inverted results and corresponding input sentences.
Our generative inversion attacks have the slightest difference in most cases.
We investigate MLC's results of SimCSE-BERT and SimCSE-RoBERTa on both datasets to see why their SWRs differ from other models.
For SimCSE-BERT on MLC, their SWRs are much smaller than the corresponding datasets' SWRs.
This is caused by several extreme cases that invert more than 10,000 tokens on embeddings of SimCSE-BERT and hence SWRs of MLC are around -20\%.
For SimCSE-RoBERTa on MLC, we find a small number of cases that output hundreds of irrelevant tokens.
Hence their SWRs are close to the datasets' SWRs by chance.

Results from both SWR and NERR confirm that previous embedding inversion attacks have little threat of breaching privacy while our GEIA can indeed recover sensitive information.

\begin{figure*}[t]
\centering
\includegraphics[width=0.999\textwidth]{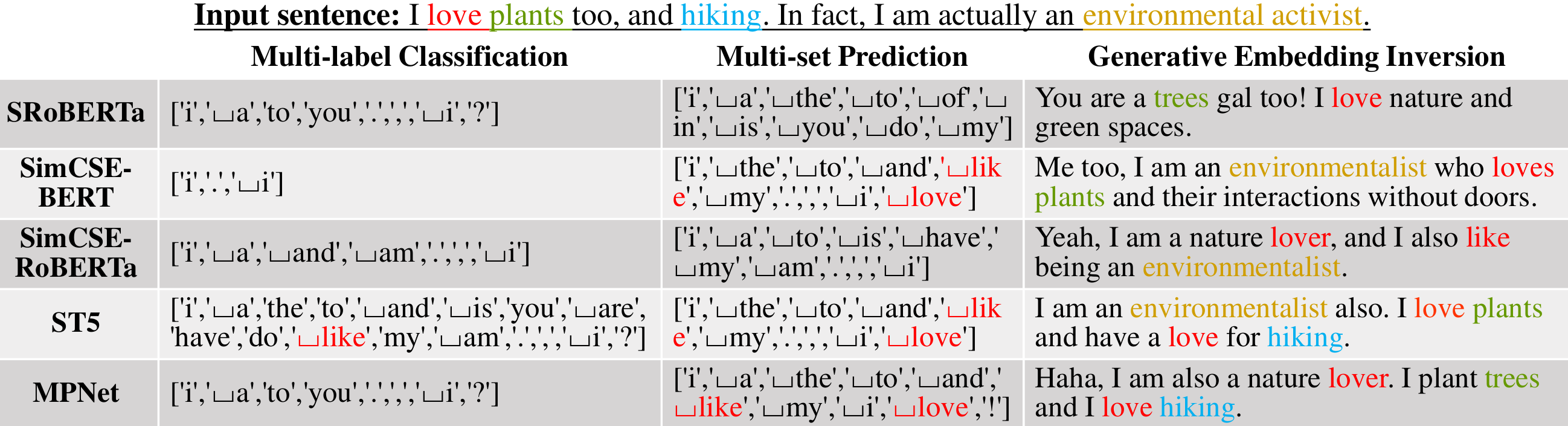}
\vspace{-0.2in}
\caption{
Embedding inversion attacks' results on the victim embedding models on the PersonaChat dataset. 
We use ``$\sqcup$'' to denote the space and highlight some informative words.
Given the same input sentence, the inverted results are shown.
Both previous embedding inversion results can only invert unordered sets of predicted tokens  while our generative embedding inversion can generate fluent sequences that are analogous to input sentences.
}
\label{fig:case}
\vspace{-0.2in}
\end{figure*}
\subsection{Evaluation on Generation}
\label{exp:Generation Quality}
Besides the classification performance and inverted tokens' sensitivity, the generation quality is also vital to generative embedding inversion attacks.
Here we view the inverted sequences as translated sentences and use generation metrics from the machine translation task to compare the similarity between inverted sequences and original inputs.
In addition, we also tune another GPT-2 model (\arr{pre-trained GPT-2\textsubscript{medium}}) with training data as the baseline.
During inference, this GPT-2 tries to decode a sentence given the first token with/without context.
Here, the context refers to all previous contents of the sentence.
For conversions of the PersonaChat dataset, the context means all the previous utterances of the current utterance in a conversation.
Our GEIA tries to invert the sentence given the embedding of this sentence.

We perform a generation evaluation on the PersonaChat dataset and the results are shown in Table~\ref{tab:generation}.
We can view BLEU-1 and ROUGE-1 as word-level precision and recall of inverted sequences, respectively.
For all 5 victim models, ROUGE-1 ranges from \arr{0.59 to 0.72} and BLEU-1 ranges from \arr{0.35 to 0.46}.
That is, \arr{more than half of the words} in the original inputs are recovered and one-third of inverted words are also in the inputs.
The results on ROUGE-1 and BLEU-1 are consistent with token-level precision and recall in Table~\ref{tab:qnli classification}.
By comparing ROUGE and BLEU with baselines, our GEIA is much more effective for inverting n-grams.
What is more, the embedding similarities of our generative inversion attacks are much higher than the GPT-2 baseline with the given context.
For fluency, we also compare the perplexity between our attacks and baselines.
Since the GPT-2 is tuned and the length of each sentence is relatively short, perplexities are small for the results.
Still, our GEIA can generate sequences with lower PPL than two baselines.
These results verify that our generative embedding attack can generate sequences well with inverted n-grams.
A more detailed evaluation of the generation is shown in the Appendix.



\subsection{Case Study}
\label{exp:case}
To show how our generative inversion attack outperforms previous baselines, we give an example of inverted results on different victim models in Figure ~\ref{fig:case}.
We manually highlight informative words in the input sentence.
If the inverted results have the same meaning as the informative words, we also highlight them using corresponding colors.
The figure shows that almost no informative word is inverted for the two baselines and the inverted tokens are mostly stop words and punctuation marks.
Therefore, it is impractical to infer the semantics for existing inversion attacks.
For our GEIA, all the inverted results can recover ``love plants'' and 3 out of 5 results can invert ``hiking'' and ``environmental activist''.
Thus, the semantics of the input sentence can be easily captured.
Despite admissible precision and recall from previous embedding inversion attacks, this example exemplifies that previous attacks cannot recover meaningful words.
On the other hand, these results justify that our generative embedding inversion attack can indeed invert informative words from sentence embeddings.
We leave more detailed contents and analyses of case studies in Appendix~\ref{app:case}.


\section{Conclusion}
In this paper, we show the limitations of existing inversion attacks and propose a generative embedding inversion attack to better recover original text sequences given their sentence embeddings.
Then we show that LM-based sentence embedding models are potentially vulnerable to our proposed attack.
We conduct extensive experiments to demonstrate the inability of previous embedding inversion attacks and disclose the privacy risks from our attack.
The defenses against the embedding inversion attack are not well studied yet, even though it is much more malicious than the attribute inference attack.
For future work, we call for more attention to effective defenses to address the embedding inversion attack with minor costs.

\section*{Limitations}
From the adversary's perspective, our attacker model's main limitation is the incapability of recovering exact domain-specific tokens.
During our experiments, we evaluate attacking results on the PersonaChat and QNLI datasets.
The PersonaChat dataset collects daily conversations between speakers with almost no expert knowledge.
The QNLI includes question-answer pairs from Wikipedia with far more domain-specific named entities than the PersonaChat dataset.
By comparing the attacking evaluations in Table~\ref{tab:qnli classification}, \ref{tab:nerr} and \ref{tab:decode methods}, all attacks on the PersonaChat dataset are more successful than attacks on the QNLI dataset.
For instance, in Table~\ref{tab:qnli classification}, F1 scores on PC are 0.1$\thicksim$0.2 larger than on QNLI on average.
In addition, QNLI 2 of Figure~\ref{app-fig-cases} shows that GEIA fails to recover the exact location ``Fresno'' 7 out of 10 times.
Though most inverted results are similar to ``What was the population of the city in 2010?'' It is hard to capture the exact city name ``Fresno''.

\section*{Ethical Considerations}
We declare that all authors of this paper acknowledge the \emph{ACM Code of Ethics} and honor the code of conduct.
This work substantially reveals potential privacy vulnerabilities of current LM-based sentence embedding models during inference.
We hereby propose the generative embedding inversion to further exploit the weaknesses of those widely used sentence embedding models.
We hope to raise more awareness of privacy leakage inside sentence embeddings and call for defenses against such information leakage.

\textbf{Data}. 
During our experiment, no personal identifiable information is used or revealed. 
Both PersonaChat and QNLI are publicly available datasets and anonymity is applied during data collection.

\textbf{Victim Embedding Models}. 
For our experiment, we use the sentence embedding models from the original GitHub repositories with given weights.
In the future, if there are other open-sourced embedding models with improved protection on privacy, we will test our proposed attack on them.

\section*{Acknowledgment}
The authors of this paper were supported by the NSFC Fund (U20B2053) from the NSFC of China, the RIF (R6020-19 and R6021-20) and the GRF (16211520 and 16205322) from RGC of Hong Kong, the MHKJFS (MHP/001/19) from ITC of Hong Kong and the National Key R\&D Program of China (2019YFE0198200) with special thanks to HKMAAC and CUSBLT. We also thank the UGC Research Matching Grants (RMGS20EG01-D, RMGS20CR11, RMGS20CR12, RMGS20EG19, RMGS20EG21, RMGS23CR05, RMGS23EG08).

\clearpage

\bibliography{anthology,custom}

\begin{thebibliography}{41}
\expandafter\ifx\csname natexlab\endcsname\relax\def\natexlab#1{#1}\fi

\bibitem[{Bojar et~al.(2016)Bojar, Chatterjee, Federmann, Graham, Haddow, Huck,
  Jimeno~Yepes, Koehn, Logacheva, Monz, Negri, Neveol, Neves, Popel, Post,
  Rubino, Scarton, Specia, Turchi, Verspoor, and
  Zampieri}]{bojar-EtAl:2016:WMT1}
Ond~{r}ej Bojar, Rajen Chatterjee, Christian Federmann, Yvette Graham, Barry
  Haddow, Matthias Huck, Antonio Jimeno~Yepes, Philipp Koehn, Varvara
  Logacheva, Christof Monz, Matteo Negri, Aurelie Neveol, Mariana Neves, Martin
  Popel, Matt Post, Raphael Rubino, Carolina Scarton, Lucia Specia, Marco
  Turchi, Karin Verspoor, and Marcos Zampieri. 2016.
\newblock \href {http://www.aclweb.org/anthology/W/W16/W16-2301} {Findings of
  the 2016 conference on machine translation}.
\newblock In \emph{Proceedings of the First Conference on Machine Translation},
  pages 131--198, Berlin, Germany. Association for Computational Linguistics.

\bibitem[{Budzianowski et~al.(2018)Budzianowski, Wen, Tseng, Casanueva, Ultes,
  Ramadan, and Ga{\v{s}}i{\'c}}]{budzianowski-etal-2018-multiwoz}
Pawe{\l} Budzianowski, Tsung-Hsien Wen, Bo-Hsiang Tseng, I{\~n}igo Casanueva,
  Stefan Ultes, Osman Ramadan, and Milica Ga{\v{s}}i{\'c}. 2018.
\newblock \href {https://doi.org/10.18653/v1/D18-1547} {{M}ulti{WOZ} - a
  large-scale multi-domain {W}izard-of-{O}z dataset for task-oriented dialogue
  modelling}.
\newblock In \emph{Proceedings of EMNLP 2018}, pages 5016--5026, Brussels,
  Belgium. Association for Computational Linguistics.

\bibitem[{Carlini et~al.(2019)Carlini, Liu, Erlingsson, Kos, and
  Song}]{Carlini-Secret-2019}
Nicholas Carlini, Chang Liu, \'{U}lfar Erlingsson, Jernej Kos, and Dawn Song.
  2019.
\newblock The secret sharer: Evaluating and testing unintended memorization in
  neural networks.
\newblock In \emph{Proceedings of USENIX Security Symposium 2019}, page
  267–284.

\bibitem[{Carlini et~al.(2021)Carlini, Tramer, Wallace, Jagielski,
  Herbert-Voss, Lee, Roberts, Brown, Song, Erlingsson, Oprea, and
  Raffel}]{carlini-2021-extracting}
Nicholas Carlini, Florian Tramer, Eric Wallace, Matthew Jagielski, Ariel
  Herbert-Voss, Katherine Lee, Adam Roberts, Tom Brown, Dawn Song, Ulfar
  Erlingsson, Alina Oprea, and Colin Raffel. 2021.
\newblock \href {https://arxiv.org/abs/2012.07805} {Extracting training data
  from large language models}.
\newblock In \emph{Proceedings of USENIX Security Symposium}, pages 2633--2650.

\bibitem[{Cer et~al.(2017)Cer, Diab, Agirre, Lopez-Gazpio, and
  Specia}]{cer-etal-2017-semeval}
Daniel Cer, Mona Diab, Eneko Agirre, I{\~n}igo Lopez-Gazpio, and Lucia Specia.
  2017.
\newblock \href {https://doi.org/10.18653/v1/S17-2001} {{S}em{E}val-2017 task
  1: Semantic textual similarity multilingual and crosslingual focused
  evaluation}.
\newblock In \emph{Proceedings of the 11th International Workshop on Semantic
  Evaluation ({S}em{E}val-2017)}, pages 1--14, Vancouver, Canada. Association
  for Computational Linguistics.

\bibitem[{Chen et~al.(2021)Chen, Chen, Yang, Lin, and Yu}]{chen2021abcd}
Derek Chen, Howard Chen, Yi~Yang, Alex Lin, and Zhou Yu. 2021.
\newblock \href {https://www.aclweb.org/anthology/2021.naacl-main.239}
  {Action-based conversations dataset: A corpus for building more in-depth
  task-oriented dialogue systems}.
\newblock In \emph{Proceedings of NAACL 2021}, pages 3002--3017, Online.
  Association for Computational Linguistics.

\bibitem[{Chung et~al.(2014)Chung, Gulcehre, Cho, and Bengio}]{GRU}
Junyoung Chung, Caglar Gulcehre, KyungHyun Cho, and Y.~Bengio. 2014.
\newblock Empirical evaluation of gated recurrent neural networks on sequence
  modeling.
\newblock In \emph{Proceedings of NIPS 2014 Deep Learning and Representation
  Learning Workshop}.

\bibitem[{Conneau and Kiela(2018)}]{conneau-kiela-2018-senteval}
Alexis Conneau and Douwe Kiela. 2018.
\newblock \href {https://aclanthology.org/L18-1269} {{S}ent{E}val: An
  evaluation toolkit for universal sentence representations}.
\newblock In \emph{Proceedings of the Eleventh International Conference on
  Language Resources and Evaluation ({LREC} 2018)}, Miyazaki, Japan. European
  Language Resources Association (ELRA).

\bibitem[{Devlin et~al.(2019)Devlin, Chang, Lee, and
  Toutanova}]{devlin-etal-2019-bert}
Jacob Devlin, Ming-Wei Chang, Kenton Lee, and Kristina Toutanova. 2019.
\newblock \href {https://doi.org/10.18653/v1/N19-1423} {{BERT}: Pre-training of
  deep bidirectional transformers for language understanding}.
\newblock In \emph{Proceedings of NAACL 2019}, pages 4171--4186, Minneapolis,
  Minnesota.

\bibitem[{Gan et~al.(2017)Gan, Pu, Henao, Li, He, and
  Carin}]{Gan-Unsupervised-2017}
Zhe Gan, Yunchen Pu, Ricardo Henao, Chunyuan Li, Xiaodong He, and Lawrence
  Carin. 2017.
\newblock \href {https://doi.org/10.18653/v1/D17-1254} {Learning generic
  sentence representations using convolutional neural networks}.
\newblock In \emph{Proceedings of EMNLP 2017}, pages 2390--2400, Copenhagen,
  Denmark. Association for Computational Linguistics.

\bibitem[{Gao et~al.(2021)Gao, Yao, and Chen}]{gao-2021-simcse}
Tianyu Gao, Xingcheng Yao, and Danqi Chen. 2021.
\newblock {SimCSE}: Simple contrastive learning of sentence embeddings.
\newblock In \emph{Proceedings of EMNLP 2021}.

\bibitem[{Gupta et~al.(2022)Gupta, Huang, Zhong, Gao, Li, and
  Chen}]{Gupta-2022-Recovering}
Samyak Gupta, Yangsibo Huang, Zexuan Zhong, Tianyu Gao, Kai Li, and Danqi Chen.
  2022.
\newblock Recovering private text in federated learning of language models.
\newblock In \emph{Proceedings of NeurIPS 2022}.

\bibitem[{Hill et~al.(2016)Hill, Cho, and Korhonen}]{hill-etal-2016-learning}
Felix Hill, Kyunghyun Cho, and Anna Korhonen. 2016.
\newblock \href {https://doi.org/10.18653/v1/N16-1162} {Learning distributed
  representations of sentences from unlabelled data}.
\newblock In \emph{Proceedings of NAACL 2016}, pages 1367--1377, San Diego,
  California. Association for Computational Linguistics.

\bibitem[{Holtzman et~al.(2020)Holtzman, Buys, Du, Forbes, and
  Choi}]{Holtzman-2020-TheCC}
Ari Holtzman, Jan Buys, Li~Du, Maxwell Forbes, and Yejin Choi. 2020.
\newblock \href {https://iclr.cc/virtual_2020/poster_rygGQyrFvH.html} {The
  curious case of neural text degeneration}.
\newblock In \emph{Proceedings of ICLR 2020}.

\bibitem[{Jang et~al.(2017)Jang, Gu, and Poole}]{Jang-Categorical-2017}
Eric Jang, Shixiang Gu, and Ben Poole. 2017.
\newblock \href {https://openreview.net/pdf?id=rkE3y85ee} {Categorical
  reparameterization with gumbel-softmax}.
\newblock In \emph{Proceedings of ICLR 2017}.

\bibitem[{Kiros et~al.(2015)Kiros, Zhu, Salakhutdinov, Zemel, Urtasun,
  Torralba, and Fidler}]{Kiros-skip-2015}
Ryan Kiros, Yukun Zhu, Russ~R Salakhutdinov, Richard Zemel, Raquel Urtasun,
  Antonio Torralba, and Sanja Fidler. 2015.
\newblock \href
  {https://proceedings.neurips.cc/paper/2015/file/f442d33fa06832082290ad8544a8da27-Paper.pdf}
  {Skip-thought vectors}.
\newblock In \emph{Proceedings of NIPS 2015}, volume~28. Curran Associates,
  Inc.

\bibitem[{Lin(2004)}]{lin-2004-rouge}
Chin-Yew Lin. 2004.
\newblock \href {https://aclanthology.org/W04-1013} {{ROUGE}: A package for
  automatic evaluation of summaries}.
\newblock In \emph{Proceedings of Text Summarization Branches Out}, pages
  74--81, Barcelona, Spain. ACL.

\bibitem[{Liu et~al.(2019)Liu, Ott, Goyal, Du, Joshi, Chen, Levy, Lewis,
  Zettlemoyer, and Stoyanov}]{liu-2019-roberta}
Yinhan Liu, Myle Ott, Naman Goyal, Jingfei Du, Mandar Joshi, Danqi Chen, Omer
  Levy, Mike Lewis, Luke Zettlemoyer, and Veselin Stoyanov. 2019.
\newblock Roberta: A robustly optimized bert pretraining approach.
\newblock \emph{arXiv preprint arXiv:1907.11692}.

\bibitem[{Ni et~al.(2022)Ni, Hernandez~Abrego, Constant, Ma, Hall, Cer, and
  Yang}]{Ni-2021-SentenceT5}
Jianmo Ni, Gustavo Hernandez~Abrego, Noah Constant, Ji~Ma, Keith Hall, Daniel
  Cer, and Yinfei Yang. 2022.
\newblock \href {https://doi.org/10.18653/v1/2022.findings-acl.146}
  {Sentence-t5: Scalable sentence encoders from pre-trained text-to-text
  models}.
\newblock In \emph{Findings of ACL 2022}, pages 1864--1874, Dublin, Ireland.
  Association for Computational Linguistics.

\bibitem[{Pan et~al.(2020)Pan, Zhang, Ji, and Yang}]{Pan-2020-Privacy}
Xudong Pan, Mi~Zhang, Shouling Ji, and Min Yang. 2020.
\newblock \href {https://doi.org/10.1109/SP40000.2020.00095} {Privacy risks of
  general-purpose language models}.
\newblock In \emph{Proceedings of 2020 IEEE Symposium on Security and Privacy
  (SP)}, pages 1314--1331.

\bibitem[{Papineni et~al.(2002)Papineni, Roukos, Ward, and
  Zhu}]{Papineni-2002-BLEU}
Kishore Papineni, Salim Roukos, Todd Ward, and Wei-Jing Zhu. 2002.
\newblock \href {https://aclanthology.org/P02-1040/} {{BLEU}: a method for
  automatic evaluation of machine translation}.
\newblock In \emph{Proceedings of ACL 2002}, pages 311--318.

\bibitem[{Qi et~al.(2020)Qi, Zhang, Zhang, Bolton, and
  Manning}]{qi-etal-2020-stanza}
Peng Qi, Yuhao Zhang, Yuhui Zhang, Jason Bolton, and Christopher~D. Manning.
  2020.
\newblock \href {https://doi.org/10.18653/v1/2020.acl-demos.14} {{S}tanza: A
  python natural language processing toolkit for many human languages}.
\newblock In \emph{Proceedings of ACL 2020: System Demonstrations}, pages
  101--108, Online. Association for Computational Linguistics.

\bibitem[{Radford et~al.(2019)Radford, Wu, Child, Luan, Amodei, and
  Sutskever}]{radford-2019-language}
Alec Radford, Jeff Wu, Rewon Child, David Luan, Dario Amodei, and Ilya
  Sutskever. 2019.
\newblock \href
  {https://d4mucfpksywv.cloudfront.net/better-language-models/language-models.pdf}
  {Language models are unsupervised multitask learners}.

\bibitem[{Raffel et~al.(2020)Raffel, Shazeer, Roberts, Lee, Narang, Matena,
  Zhou, Li, and Liu}]{2020t5}
Colin Raffel, Noam Shazeer, Adam Roberts, Katherine Lee, Sharan Narang, Michael
  Matena, Yanqi Zhou, Wei Li, and Peter~J. Liu. 2020.
\newblock \href {http://jmlr.org/papers/v21/20-074.html} {Exploring the limits
  of transfer learning with a unified text-to-text transformer}.
\newblock \emph{Journal of Machine Learning Research}, 21(140):1--67.

\bibitem[{Rajpurkar et~al.(2016)Rajpurkar, Zhang, Lopyrev, and
  Liang}]{rajpurkar-etal-2016-squad}
Pranav Rajpurkar, Jian Zhang, Konstantin Lopyrev, and Percy Liang. 2016.
\newblock \href {https://doi.org/10.18653/v1/D16-1264} {{SQ}u{AD}: 100,000+
  questions for machine comprehension of text}.
\newblock In \emph{Proceedings of EMNLP 2016}, pages 2383--2392. Association
  for Computational Linguistics.

\bibitem[{Reimers and Gurevych(2019)}]{reimers-2019-sentence-bert}
Nils Reimers and Iryna Gurevych. 2019.
\newblock \href {https://arxiv.org/abs/1908.10084} {Sentence-bert: Sentence
  embeddings using siamese bert-networks}.
\newblock In \emph{Proceedings of EMNLP 2019}. Association for Computational
  Linguistics.

\bibitem[{Sennrich et~al.(2016)Sennrich, Haddow, and
  Birch}]{sennrich-etal-2016-neural}
Rico Sennrich, Barry Haddow, and Alexandra Birch. 2016.
\newblock \href {https://doi.org/10.18653/v1/P16-1162} {Neural machine
  translation of rare words with subword units}.
\newblock In \emph{Proceedings of ACL 2016}, pages 1715--1725.

\bibitem[{Socher et~al.(2013)Socher, Perelygin, Wu, Chuang, Manning, Ng, and
  Potts}]{socher-etal-2013-recursive}
Richard Socher, Alex Perelygin, Jean Wu, Jason Chuang, Christopher~D. Manning,
  Andrew Ng, and Christopher Potts. 2013.
\newblock \href {https://www.aclweb.org/anthology/D13-1170} {Recursive deep
  models for semantic compositionality over a sentiment treebank}.
\newblock In \emph{Proceedings of EMNLP 2013}, pages 1631--1642, Seattle,
  Washington, USA. Association for Computational Linguistics.

\bibitem[{Song and Raghunathan(2020)}]{song-information-2020}
Congzheng Song and Ananth Raghunathan. 2020.
\newblock \href {https://doi.org/10.1145/3372297.3417270} {Information leakage
  in embedding models}.
\newblock In \emph{Proceedings of ACM CCS 2020}, page 377–390.

\bibitem[{Song et~al.(2017)Song, Ristenpart, and Shmatikov}]{Song2017MachineLM}
Congzheng Song, Thomas Ristenpart, and Vitaly Shmatikov. 2017.
\newblock Machine learning models that remember too much.
\newblock \emph{Proceedings of the 2017 ACM SIGSAC Conference on Computer and
  Communications Security}.

\bibitem[{Song et~al.(2020)Song, Tan, Qin, Lu, and Liu}]{song-2020-mpnet}
Kaitao Song, Xu~Tan, Tao Qin, Jianfeng Lu, and Tie-Yan Liu. 2020.
\newblock \href
  {https://proceedings.neurips.cc/paper/2020/file/c3a690be93aa602ee2dc0ccab5b7b67e-Paper.pdf}
  {Mpnet: Masked and permuted pre-training for language understanding}.
\newblock In \emph{Proceedings of the NeurIPS 2020}.

\bibitem[{Thakkar et~al.(2021)Thakkar, Ramaswamy, Mathews, and
  Beaufays}]{thakkar-2021-understanding}
Om~Dipakbhai Thakkar, Swaroop Ramaswamy, Rajiv Mathews, and Francoise Beaufays.
  2021.
\newblock \href {https://doi.org/10.18653/v1/2021.privatenlp-1.1}
  {Understanding unintended memorization in language models under federated
  learning}.
\newblock In \emph{Proceedings of the Third Workshop on Privacy in Natural
  Language Processing}, pages 1--10, Online. Association for Computational
  Linguistics.

\bibitem[{Wang et~al.(2018)Wang, Singh, Michael, Hill, Levy, and
  Bowman}]{wang-etal-2018-glue}
Alex Wang, Amanpreet Singh, Julian Michael, Felix Hill, Omer Levy, and Samuel
  Bowman. 2018.
\newblock \href {https://doi.org/10.18653/v1/W18-5446} {{GLUE}: A multi-task
  benchmark and analysis platform for natural language understanding}.
\newblock In \emph{Proceedings of the 2018 {EMNLP} Workshop {B}lackbox{NLP}:
  Analyzing and Interpreting Neural Networks for {NLP}}, pages 353--355.

\bibitem[{Welleck et~al.(2018)Welleck, Yao, Gai, Mao, Zhang, and
  Cho}]{Welleck-2018-Multiset}
Sean Welleck, Zixin Yao, Yu~Gai, Jialin Mao, Zheng Zhang, and Kyunghyun Cho.
  2018.
\newblock \href
  {https://proceedings.neurips.cc/paper/2018/file/fb3f76858cb38e5b7fd113e0bc1c0721-Paper.pdf}
  {Loss functions for multiset prediction}.
\newblock In \emph{Proceedings of NIPS 2018}, volume~31. Curran Associates,
  Inc.

\bibitem[{Williams et~al.(2018)Williams, Nangia, and
  Bowman}]{Williams-mnli-2018}
Adina Williams, Nikita Nangia, and Samuel Bowman. 2018.
\newblock \href {http://aclweb.org/anthology/N18-1101} {A broad-coverage
  challenge corpus for sentence understanding through inference}.
\newblock In \emph{Proceedings of NAACL 2018}, pages 1112--1122. Association
  for Computational Linguistics.

\bibitem[{Williams and Zipser(1989)}]{Williams-1989-teacher}
Ronald~J. Williams and David Zipser. 1989.
\newblock A learning algorithm for continually running fully recurrent neural
  networks.
\newblock \emph{Neural Computation}, 1(2):270--280.

\bibitem[{Zhang et~al.(2018{\natexlab{a}})Zhang, Wu, Li, and
  Li}]{zhang-etal-2018-learning-universal}
Minghua Zhang, Yunfang Wu, Weikang Li, and Wei Li. 2018{\natexlab{a}}.
\newblock \href {https://doi.org/10.18653/v1/D18-1481} {Learning universal
  sentence representations with mean-max attention autoencoder}.
\newblock In \emph{Proceedings of EMNLP 2018}, pages 4514--4523, Brussels,
  Belgium. Association for Computational Linguistics.

\bibitem[{Zhang et~al.(2018{\natexlab{b}})Zhang, Dinan, Urbanek, Szlam, Kiela,
  and Weston}]{zhang-etal-2018-personalizing}
Saizheng Zhang, Emily Dinan, Jack Urbanek, Arthur Szlam, Douwe Kiela, and Jason
  Weston. 2018{\natexlab{b}}.
\newblock \href {https://doi.org/10.18653/v1/P18-1205} {Personalizing dialogue
  agents: {I} have a dog, do you have pets too?}
\newblock In \emph{Proceedings of ACL 2018}, pages 2204--2213.

\bibitem[{Zhang et~al.(2022)Zhang, Roller, Goyal, Artetxe, Chen, Chen, Dewan,
  Diab, Li, Lin, Mihaylov, Ott, Shleifer, Shuster, Simig, Koura, Sridhar, Wang,
  and Zettlemoyer}]{OPT}
Susan Zhang, Stephen Roller, Naman Goyal, Mikel Artetxe, Moya Chen, Shuohui
  Chen, Christopher Dewan, Mona Diab, Xian Li, Xi~Victoria Lin, Todor Mihaylov,
  Myle Ott, Sam Shleifer, Kurt Shuster, Daniel Simig, Punit~Singh Koura, Anjali
  Sridhar, Tianlu Wang, and Luke Zettlemoyer. 2022.
\newblock \href {https://doi.org/10.48550/ARXIV.2205.01068} {Opt: Open
  pre-trained transformer language models}.

\bibitem[{Zhao et~al.(2020)Zhao, Mopuri, and Bilen}]{Zhao-2020-iDLG}
Bo~Zhao, Konda~Reddy Mopuri, and Hakan Bilen. 2020.
\newblock idlg: Improved deep leakage from gradients.
\newblock volume abs/2001.02610.

\bibitem[{Zhu et~al.(2019)Zhu, Liu, and Han}]{Zhu-2019-Deep}
Ligeng Zhu, Zhijian Liu, and Song Han. 2019.
\newblock \href
  {https://proceedings.neurips.cc/paper/2019/file/60a6c4002cc7b29142def8871531281a-Paper.pdf}
  {Deep leakage from gradients}.
\newblock In \emph{Proceedings of NIPS 2019}, volume~32. Curran Associates,
  Inc.

\end{thebibliography}
\bibliographystyle{acl_natbib}

\clearpage
\appendix

\section{Training Details}
\arr{\textbf{Hyper-parameters and setups}.} For the multi-label classification, we use a 1-layer neural network as the attacker. 
To determine the thresholds for classification, we perform the grid search with an interval of 0.05.
The multi-set prediction uses a unidirectional GRU~\cite{GRU} of 10 time steps as the attacker.
For every time step, we use sentence embedding as the input with the multi-set objective~\cite{Welleck-2018-Multiset}. 
We experiment with different time steps and inputs (e.g., averaging the sentence embedding with corresponding token embedding for a time step) and find that the time step of 10 with only sentence embeddings yields the best performance.
The baselines and our GEIA use the same byte pair encoding tokenizer~\cite{sennrich-etal-2016-neural} for a fair comparison.
Our GEIA uses a \arr{randomly initialized} GPT-2 medium model (345M) \cite{radford-2019-language}.
The decoding uses the beam search with beam size 5.
For all 3 models, we use the Adam optimizer to update the models with a learning rate of 3e-4 \arr{and batch size of 64}.

\arr{\textbf{Training details}. During training, we first obtain batches of sentence embeddings from victim embedding models and project embeddings to the exact dimension of the attacker's token representations.
Then we gather the corresponding batches of tokens' representations by passing tokens through the attacker's embedding layers with attention masks.
As shown in Figure~\ref{fig:model}, we concatenate sentence embeddings to the left of the tokens' representations as the inputs.
Hence, the sentence embeddings can be viewed as the initial token representations followed by the original tokens' representations.
Lastly, we feed concatenated representations to train the attacker with the language modeling objective.}

\arr{Take the sentence $x=$``$w_0 w_1 ... w_{u-1}$'' as one example, we use $Align(f(x))$ to denote the aligned sentence embedding and $\Phi_{emb}(w_i)$ to denote the representation of token $w_i$ of attacker model $\Phi$.
Our input $I$ is \text{
\footnotesize{
$[Align(f(x)), \Phi_{emb}(w_0),\Phi_{emb}(w_1),...,\Phi_{emb}(w_{u-1})]$}
}.
And our prediction manages to maximize the probability of the target sequence $O = $ [$w_0,w_1,...,w_{u-1},$<eos>], where the <eos> indicates the special end of sentence token.
Both $I$ and $O$ are of length $u+1$.
For each time step $t$ where $0<t<u$, our attacker aims to output $O_t=w_t$ given $Align(f(x))$, $\Phi_{emb}(w_0),...,\Phi_{emb}(w_{t-1})$.
If $t=0$, the desired output is $O_0=w_0$ given only the sentence embedding $Align(f(x))$.
If $t=u$, the desired output is $O_u=$ <eos> given the whole input sequence $I$.
}

\textbf{Toolkits}. 
For finding named entities, we use Stanza toolkit~\cite{qi-etal-2020-stanza} to extract named entities from two datasets.
We use the NLTK package to measure BLEU scores and Huggingface's Evaluate library to measure ROUGE scores.
For micro-averaged scores, we use the sklearn library to calculate precision, recall and F1.

During our experiment, we use 2 NVIDIA GeForce RTX 3090 to run our codes and it takes around 7 hours to train the attacker of GEIA for 10 epochs.

\section{Details of Victim Embedding Models}
\label{sec:Details of Victim}
In this section, we give more details of victim embedding models used in our experiments and their checkpoints.

\arr{
$\bullet$ Sentence-RoBERTa (SRoBERTa) ~\cite{reimers-2019-sentence-bert}: Sentence-BERT proposes a siamese network to reduce computational overhead for sentence embeddings. 
We adopt the Sentence-RoBERTa model since it has better performance than SBERT.
In our experiment, ``all-roberta-large-v1'' (355M) is used as SRoBERTa.}

$\bullet$ SimCSE ~\cite{gao-2021-simcse}: SimCSE considers the simple contrastive learning objective by self-prediction with dropout. And SimCSE performs better than SBERT and SRoBERTa.
For our experiment, we use both SimCSE-BERT (``sup-simcse-bert-large-uncased,'' 340M) and SimCSE-RoBERTa (``sup-simcse-roberta-large,'' 355M) as victim models.

$\bullet$ Sentence-T5 (ST5) ~\cite{Ni-2021-SentenceT5}: Sentence-T5 exploits the encoder of T5 model architecture ~\cite{2020t5} to achieve the new state-of-the-art on sentence embedding tasks. In our experiment, ``sentence-t5-large'' (770M) is used.

$\bullet$ MPNet ~\cite{song-2020-mpnet}: MPNet proposed a  unified learning objective for BERT to combine masked language modeling and permuted language modeling.
In our experiment, ``all-mpnet-base-v1'' (110M) is used.


\begin{table*}[t]
\centering
\small
  \begin{tabular}{c l  c cc ccc  ccc }
    \toprule
    \multirow{2}{*}{Attacker} &
    \multirow{2}{*}{Victim} &
    \multirow{2}{*}{ES} &
      \multicolumn{2}{c}{ROUGE} &
      \multicolumn{2}{c}{BLEU} &
      \multicolumn{3}{c}{Classification} &
      \multicolumn{1}{c}{\multirow{2}{*}{EMR}} 
       \\
      {} & {} & {} & {ROUGE-1} & {ROUGE-L} & {BLEU-1} & {BLEU-4} & Pre & Rec & F1 & {}\\
      \midrule
      \multirow{4}{*}{GPT-2\textsubscript{345M}} 
     & ST5\textsubscript{220M}             & 91.08 & 66.62 & 61.19 & 43.52 & 19.88 & 59.86 & 56.11 & 57.93 & 17.79 \\ 
     & ST5\textsubscript{770M}            & 90.26 & 62.18 & 57.11 & 40.10 & 17.53 & 57.20 & 51.69 & 54.31 & 14.48 \\ 
     & ST5\textsubscript{3B}             & 90.93 & 64.36 & 59.26 & 41.86 & 18.53 & 57.86 & 54.28 & 56.01 & 16.97 \\ 
     & ST5\textsubscript{11B}            & 91.15 & 63.98 & 59.04 & 41.86 & 18.35 & 57.53 & 54.16 & 55.79 & 16.66 \\ 
     
     \midrule
     \midrule
      \multirow{4}{*}{GPT-2\textsubscript{762M}} 
     & ST5\textsubscript{220M}             & 91.61 & 68.66 & 63.51 & 43.99 & 21.07 & 62.71 & 57.21 & 59.83 & 23.69 \\ 
     & ST5\textsubscript{770M}            & 90.96 & 65.55 & 60.97 & 42.09 & 19.62 & 59.74 & 54.92 & 57.22 & 18.00 \\ 
     & ST5\textsubscript{3B}                & 91.44 & 66.04 & 61.37 & 42.25 & 19.77 & 60.35 & 55.19 & 57.66 & 18.08 \\ 
     & ST5\textsubscript{11B}               & 90.43 & 62.14 & 57.53 & 40.35 & 17.56 & 56.10 & 52.29 & 54.12 & 16.03 \\ 

    \bottomrule
  \end{tabular}
  \vspace{-0.05in}
  \caption{\label{tab:model size}
Generative embedding inversion attacks' performance on ST5 with the different victim and attacker sizes.
}
\vspace{-0.05in}
\end{table*}


\begin{table*}[t]
\centering
\small
  \begin{tabular}{l  c cc ccc  ccc }
    \toprule
    \multirow{2}{*}{Attacker} &
    \multirow{2}{*}{ES} &
      \multicolumn{2}{c}{ROUGE} &
      \multicolumn{2}{c}{BLEU} &
      \multicolumn{3}{c}{Classification} &
      \multicolumn{1}{c}{\multirow{2}{*}{EMR}} 
       \\
      {}  & {} & {ROUGE-1} & {ROUGE-L} & {BLEU-1} & {BLEU-4} & Pre & Rec & F1 & {}\\
      \midrule

     GPT-2\textsubscript{345M}               
     & 90.26 & 62.18 & 57.11 & 40.10 & 17.53 & 57.20 & 51.69 & 54.31 & 14.48  \\ 
     GPT-2\textsubscript{345M-random}               
     & 91.47 & 70.72 & 65.45 & 44.52 & 21.99 & 67.46 & 58.26 & 62.53 & 19.11 \\
     GPT-2\textsubscript{762M} 
     & 90.96 & 65.55 & 60.97 & 42.09 & 19.62 & 59.74 & 54.92 & 57.22 & 18.00 \\
     OPT\textsubscript{350m} 
     & 90.04 & 62.86 & 58.56 & 39.24 & 18.07 & 58.47 & 51.97 & 55.02 & 16.49 \\ 
     OPT\textsubscript{1.3b} 
     & 94.12 & 74.39 & 69.40 & 47.70 & 25.11 & 68.40 & 62.27 & 65.19 & 23.79 \\ 
     T5\textsubscript{770M-lm-adapt} 
     & 90.57 & 63.76 & 58.62 & 43.20 & 18.75 & 54.96 & 54.92 & 54.94 & 17.01 \\ 
     T5\textsubscript{770M-random}   
     & 81.35 & 46.16 & 43.27 & 27.05 & 9.58 & 47.31 & 37.64 & 41.92 &  6.92 \\ 

    \bottomrule
  \end{tabular}
  \vspace{-0.05in}
  \caption{\label{tab:decoder setups}
\arr{Generative embedding inversion attacks' performance on ST5\textsubscript{large} (770M) with the different attacker models and initializations.The results are evaluated on the PersonaChat dataset.}
}
\vspace{-0.05in}
\end{table*}

\section{Evaluations on Exact Match Ratio and Edit Distance}
\label{app:Exact Match}
In addition to the similarity, we are also interested in how many inverted sequences are verbatim input sentences.
If they are not the same, we would like to know the minimal edits to modify the inverted sequence to the inputs.
In this part, we use the exact match ratio (EMR) to calculate the ratio of inverted sequences that are exactly the same as inputs after removing punctuation.
We also report the mean and median of edit distance (ED).
The edit distance, also known as Levenshtein distance, measures the minimal changes of characters needed to update the inverted sequence to the actual input.

The evaluation results are shown in Table ~\ref{tab:decode methods}.
For EMR, our GEIA can recover approximately 10\% of verbatim sentences on the PersonaChat dataset.
However, our GEIA inverts no more than 1\% exact sequences on the QNLI dataset.
For edit distance, GEIA needs around 28 edits on the PersonaChat dataset and 85 edits on the QNLI dataset.
These results show that our GEIA can indeed recover verbatim input sentences from their embeddings.
Still, GEIA cannot handle domain knowledge well and the performance drops on the QNLI dataset.

\section{Evaluation on Decoding Methods}
\label{app:decode}
Our experiments implement beam search decoding for the generation process.
In this section, we compare beam search decoding with sampling-based decoding.
We use the Nucleus Sampling ~\cite{Holtzman-2020-TheCC} method to sample tokens.
We set top-p = 0.9 with a temperature
coefficient 0.9.

The evaluation results of the two decoding methods are shown in Table ~\ref{tab:decode methods}.
Both generation and classification metrics are included in the two datasets.
Except for a few results of BLEU-1 and Recall on the QNLI dataset, beam search significantly outperforms Nucleus Sampling on various metrics.
For example, compared with Nucleus Sampling, beam search brings 3\% - 5\% improvements on the F1 and 1\% - 3\% improvements on the BLEU-4.
Additionally, beam search leads to higher EMRs and smaller edits.
These results help explain why we use the beam search decoding for our experiments.

\section{Evaluations on Models' Sizes}
\label{app:model size}
In this section, to study the attack performance on LM-based embedding models of different scales, we perform generative embedding inversion attacks on a specific victim model with different model sizes and attacker sizes.

In Table~\ref{tab:model size}, we evaluate GEIA on pre-trained ST5 of four different model sizes from ST5\textsubscript{base} (220M) to ST5\textsubscript{xxl} (11B) on the PersonaChat dataset.
We use GPT-2\textsubscript{medium} (345M) and GPT-2\textsubscript{large} (762M) as attackers.
The good attacking results suggest that GEIA is still effective despite different model scales.
By comparing attacking performance on the same victim model between GPT-2\textsubscript{medium} and GPT-2\textsubscript{large}, we found that embedding models are generally more vulnerable after increasing the attacker's capacity.
In addition, we find that ST5\textsubscript{base} tends to be the most vulnerable to GEIA while other models are more robust towards GEIA.
This suggests that small-sized embedding models are more unsafe towards GEIA than the larger models.

\section{Evaluations on Different Attacker Models and Initializations}
\label{app:attacker size}
\arr{To show that GEIA can be adaptive with various powerful decoders, we evaluate the performance of GEIA with different attacker models and initializations on the PersonaChat dataset with ST5\textsubscript{large} (770M) as the victim model.
Besides GPT-2, we extend attacker models to OPT~\cite{OPT} and T5 and perform GEIA on ST5\textsubscript{large}.
GPT-2 and OPT are built on the transformer's decoder blocks and pre-trained with different datasets while T5 consists of both encoder and decoder blocks.
For GEIA with T5 as the attacker, we feed sentence embeddings to the encoder and perform decoding on the decoder side.
For GPT-2, we use randomly initialized GPT-2\textsubscript{medium-random} (345M), pre-trained GPT-2\textsubscript{medium} (345M) and pre-trained GPT-2\textsubscript{large} (762M) as attackers.
For OPT, we experiment with pre-trained OPT\textsubscript{350m} and OPT\textsubscript{1.3b}.
For T5, we evaluate performance on randomly initialized T5\textsubscript{large-random} (770M) and LM-adapted T5\textsubscript{large-lm-adapt}\footnote{\url{https://huggingface.co/google/t5-large-lm-adapt}.} (770M).
}

\arr{In Table~\ref{tab:decoder setups}, we can see that all our attackers can outperform the previous baselines of Table~\ref{tab:qnli classification} on classification.
Interestingly, our results suggest that pre-training does not constantly improve performance than random initialization for GEIA.
In terms of beam search, Pre-trained T5\textsubscript{large-lm-adapt} outperforms T5\textsubscript{large-random} while GPT-2\textsubscript{medium-random} defeats both  GPT-2\textsubscript{medium} and GPT-2\textsubscript{large}.
Since our GEIA has a different training paradigm from LMs' pre-training objectives, it is hard to conclude whether pre-training can improve GEIA or not.
However, pre-training helps attackers better model the probability distribution of tokens and enhances the performance of sampling-based decoding strategies.
In addition, our results on model scales of GPT-2 and OPT help verify that GEIA can be improved by increasing attackers' sizes.
}

\section{More on Case Studies}
\label{app:case}
\begin{figure*}
    \centering  
	\subfigcapskip=-5pt 
	\subfigure{
	
		\includegraphics[width=0.999\textwidth]{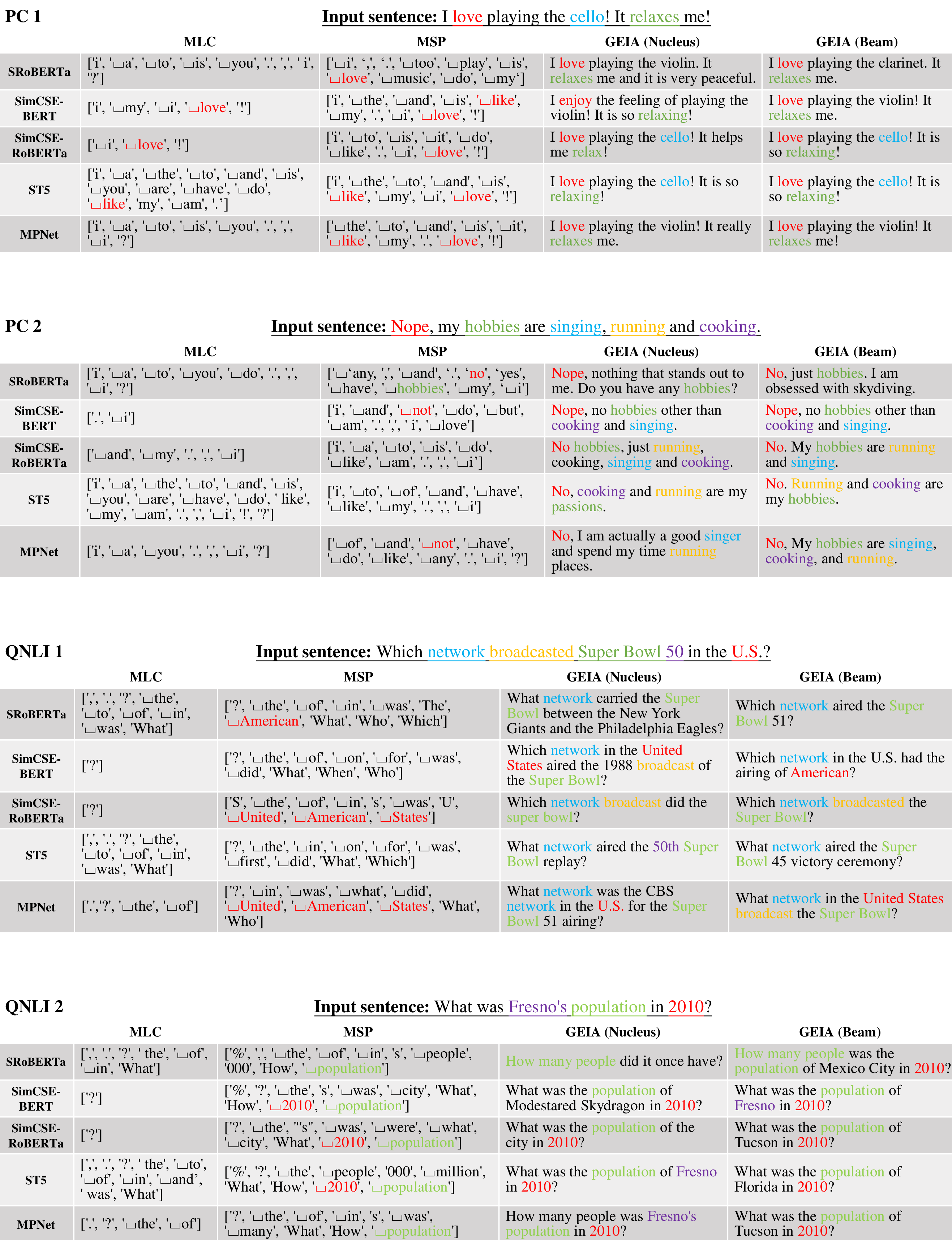}}
	
	  
	
	\caption{\label{app-fig-cases}More cases of embedding inversion attacks.}
\end{figure*}

In this section, we give more cases on two datasets to show the effectiveness of GEIA intuitively.
Figure~\ref{app-fig-cases} gives two examples for each dataset with 2 decoding methods included.
Still, the informative words are highlighted manually for both input sentences and inverted results.
For all cases, MLC performs the worst: only the token ``love'' is inverted 3 times on the first example of PC.
And the remained inverted results are mostly meaningless stop words.
MSP performs better than MLC: some informative words like ``US'' and ``population'' can be recovered on the QNLI dataset.
For GEIA, both decoding algorithms can recover many relevant words and generate coherent sentences.
Moreover, some digits can even be mined: the year ``2010'' of QNLI 2 is successfully recovered for 9 out of 10 cases.
On QNLI 1, the number ``50'' is also inverted by GEIA on ST5.
Interestingly, both GEIA also try to predict the numbers during generation: beam search decoding predicts ``45'' and ``51'' while  Nucleus Sampling outputs ``51.''
This example suggests that GEIA is also aware of digits like years.
On PC 2, GEIA can even capture all 3 hobbies and invert them correctly.

In summary, these cases show that previous MLC and MSP perform poorly on embedding inversion and our GEIA works much better than previous works.

\begin{table*}[t]
\centering
\small

\begin{tabular}{l l c c c c}
\hline
{Name} & \textbf{Task}  & \textbf{Sentences \#} & \textbf{Train/dev/test} & \textbf{Unique NEs}  & \textbf{Avg. Sentence Len}\\
\hline
ABCD            & Goal-oriented dialogues  & 184,501 & 80:10:10 & 7,306 & 8.18 \\
MNLI            & Natural Language Inference (NLI) & 824,626 & 95.1:2.4:2.4 & 31,990 & 14.88 \\
MultiWOZ        & Intent tracking, dialog prediction & 143,044 & 80:10:10 & 18,971 & 13.23 \\
SST-2           & Sentiment Analysis (SA)  & 70,042 & 96.2:1.2:2.6 & 758 & 9.79 \\
WMT16           & Machine Translation (MT) & 1,002,895 & 99.4:0.3:0.3 & 8,904 & 23.19 \\

\hline
\end{tabular}

\vspace{-0.05in}
\caption{\label{app-tab:dataset-table}
Statistics of datasets.
}
\vspace{-0.1in}
\end{table*}

\begin{table*}[t]
\centering
\small
  \begin{tabular}[t]{l  cccc | ccc | ccc}
    \toprule

    \multirow{2}{*}{Dataset} &
      \multicolumn{4}{c|}{MLC} &
      \multicolumn{3}{c|}{MSP} &
      \multicolumn{3}{c}{GEIA} \\
      {} & {Threshold} & {Pre} & {Rec} & {F1} & {Pre} & {Rec} & {F1} & {Pre} & {Rec} & {F1}  \\
      \midrule

     ABCD              & 0.45 & 80.18 & 36.61 & 50.26  & 60.23 & 56.59 & 58.35 & \textbf{84.21} & \textbf{77.42} & \textbf{80.67} \\
     MNLI              & 0.50 & \textbf{80.56} & 20.93 & 33.23  & 59.64 & 34.33 & 43.57 & 59.51 & \textbf{44.72} & \textbf{51.07} \\
     MultiWOZ          & 0.50 & 82.65 & 34.22 & 48.40  & 79.37 & 46.47 & 58.62 & \textbf{86.47} & \textbf{77.88} & \textbf{81.95} \\
     SST-2             & 0.75 & 50.31 & 4.92 & 8.96    & \textbf{52.95} & \textbf{23.95} & \textbf{32.98} & 35.06 & 11.99 & 17.87 \\
     WMT16             & 0.70 & \textbf{81.86} & 13.16 & 22.68 & 55.56 & 24.61 & 34.11  & 46.16 & \textbf{31.79} & \textbf{37.65} \\

    \bottomrule
  \end{tabular}
  \vspace{-0.05in}
  \caption{
  \label{app-tab:more datasets}
Embedding inversion performance on classification metrics. 
The evaluations are done on the embeddings of SimCSE-BERT.
Precision (Pre), recall (Rec) and F1 are measured in \%.
}
\vspace{-0.1in}
\end{table*}
\begin{table*}[!htbp]
\centering
\small
  \begin{tabular}[t]{l  cccc | ccc}
    \toprule
    \multirow{2}{*}{Victim Model} &
      \multicolumn{4}{c|}{SWR} &
      \multicolumn{3}{c}{NERR}  \\
     {} & {Test set} & {MLC} & {MSP} & {GEIA} & {MLC} & {MSP} & {GEIA}   \\
      \midrule

     ABCD               & 39.74 & +09.85 & +06.60 & \textbf{-01.42} & 14.01 & 23.66 & \textbf{52.97} \\
     MNLI               & 42.66 & +23.71 & +21.32 & \textbf{+00.57} & 02.19 & 05.04 & \textbf{33.93} \\
     MultiWOZ           & 38.92 & +15.85 & +09.23 & \textbf{-00.19} & 06.16 & 07.98 & \textbf{60.67} \\
     SST-2              & 48.17 & -42.93 & +20.76 & \textbf{+19.00} & 00.00 & \textbf{03.42} & 00.79 \\
     WMT16              & 40.22 & +24.80 & +34.97 & \textbf{+04.24} & 00.99 & 01.80 & \textbf{18.91} \\

    \bottomrule
  \end{tabular}
  \vspace{-0.05in}
  \caption{\label{app-tab:nerr}
Embedding inversion performance on stop word rate (SWR) and named entity recovery ratio (NERR). 
All attacks are conducted on SimCSE-BERT.
NERR and SWR are measured in \%.
For SWR, we report the SWRs of testing data as baselines and the differences between baselines' SWRs and SWRs of various embedding inversion attacks.
}
\vspace{-0.1in}
\end{table*}

\section{GEIA on More Datasets}
\label{app:more data}
To demonstrate that our proposed GEIA is universal and can be applied to any textual data, here we evaluate GEIA on 5 more datasets of different domains, tasks and scales.
Without loss of generality, we set SimCSE-BERT as the victim model and perform inversion attacks on Action-Based Conversations Dataset (ABCD) \cite{chen2021abcd}, Multi-Genre Natural Language Inference (MNLI) \cite{Williams-mnli-2018}, Multi-domain Wizard-of-Oz (MultiWOZ) \cite{budzianowski-etal-2018-multiwoz}, Stanford Sentiment Treebank v2 (SST2) \cite{socher-etal-2013-recursive} and WMT16 \cite{bojar-EtAl:2016:WMT1}.
A detailed summary statistics of these five datasets is shown in Table~\ref{app-tab:dataset-table}.

Table~\ref{app-tab:more datasets} and \ref{app-tab:nerr} compares GEIA with previous baselines in classification and informativeness, separately.
We can still observe that our GEIA can mostly outperform MLC and MSP with better recall and F1 on both classification and informativeness.
One exception is that MSP outperforms GEIA on SST-2.
We examined the inverted contents of GIEA on SST-2, and found that GIEA frequently generated repetitions of meaningless stop words due to insufficient training data (merely 67,349 sentences).
The randomly initialized GPT-2 cannot generalize well on this small-scale dataset and therefore performs poorly on embedding inversion.

After inspecting the attacking performance on all seven datasets, including PC and QNLI, we discovered that embedding inversion attacks' performance is dependent on data scales, domains and informativeness of contents.
Still, given a reasonable amount of training data, our GEIA can easily exceed previous baselines on both classification and informativeness.
Moreover, our proposed GEIA changes the previous classification paradigm to generation and can recover ordered sequences.

\newpage

\begin{sidewaystable*}
\centering
\small
  \begin{tabular}{c l c  c c cc ccc ccc c cc}
    \toprule
    \multirow{2}{*}{Data} &
    \multirow{2}{*}{Victim Model} &
    \multicolumn{1}{c}{Initialization} &
    \multicolumn{1}{c}{\multirow{2}{*}{PPL}} &
    \multicolumn{1}{c}{\multirow{2}{*}{ES}} &
      \multicolumn{2}{c}{ROUGE} &
      \multicolumn{3}{c}{BLEU} &
      \multicolumn{3}{c}{Classification} &
      \multicolumn{1}{c}{\multirow{2}{*}{EMR}} &
      \multicolumn{2}{c}{ED}
       \\
       \cmidrule(lr){3-3}
      {} & {} & {Decode} & {} & {} & {ROUGE-1} & {ROUGE-L} & {BLEU-1} & {BLEU-2} & {BLEU-4} & Pre & Rec & F1 & {} & Mean & Median \\
      \midrule
      \multirow{20}{*}{PC} 
    & SRoBERTa        & & 7.52 & 83.77 & 44.55 & 41.00 & 30.16 & 13.31 & 9.75 & 39.60 & 39.54 & 39.57 & 8.71 & 33.10 & 35\\
    & SimCSE-BERT     & GPT-2\textsubscript{pretrain}
    & 8.70 & 86.56 & 54.83 & 48.94 & 38.24 & 19.01 & 14.15 & 46.81 & 47.24 & 47.03  & 11.05 & 31.16 & 32 \\
    & SimCSE-RoBERTa  
    & ------------------
    & 8.46 & 86.48 & 50.56 & 44.90 & 35.20 & 16.04 & 11.53 & 44.09 & 44.13 & 44.11 & 8.43 & 32.63 & 34 \\
    & ST5             & Nucleus
    & 8.45 & 88.31 & 55.55 &  50.21 & 38.42 & 19.18 & 14.29 & 48.42 & 48.32 & 48.37 & 11.54 & 29.83 & 31 \\
    & MPNet           & & 8.30 & 85.88 & 50.49 & 45.88 & 34.53 & 16.21 & 12.03 & 44.32 & 44.31 & 44.32 & 10.44 & 31.45 & 33 \\
    
    \cmidrule(lr){2-16}
    & SRoBERTa        & 
    & 4.78 & 85.81 & 50.68 & 47.30 & 31.62 & 16.19 & 12.22 & 47.92 & 42.65 & 45.13 & 10.40 & 29.63 & 30 \\
    & SimCSE-BERT     & GPT-2\textsubscript{pretrain}
    & 6.68 & 88.23 & 60.02 & 54.09 & 41.02 & 22.25 & 16.85 & 52.52 & 50.88 & 51.69 & 12.99 & 28.43 & 29 \\
    & SimCSE-RoBERTa  
    & ------------------ 
    & 5.33 & 88.19 & 56.45 & 51.10 & 36.22 & 18.84 & 14.01 & 51.86 & 46.78 & 49.19 & 10.43 & 28.95 & 30 \\
    & ST5             & Beam
    & 5.50 & 90.26 & 62.18 & 57.11 & 40.10 & 22.84 & 17.53 & 57.20 & 51.69 & 54.31 & 14.48 & 26.15 & 26 \\
    & MPNet           & & 5.39 & 87.89 & 56.82 & 52.38 & 36.21 & 19.56 & 14.90 & 52.78 & 47.68 & 50.10 & 12.69 & 27.87 & 28 \\
    
    \cmidrule(lr){2-16}
    & SRoBERTa         & 
    & 9.27 & 86.53 & 55.03 & 51.05 & 36.75 & 18.65 & 13.89 & 49.58 & 47.87 & 48.71 & 11.80 & 28.53 & 29 \\
    & SimCSE-BERT     & GPT-2\textsubscript{random}
    & 10.14 & 89.94 & 68.20 & 60.99 & 47.23 & 26.95 & 20.81 & 59.70 & 58.23 & 58.95 & 17.24 & 25.49 & 25 \\
    & SimCSE-RoBERTa  
    & ------------------ 
    & 9.71 & 90.02 & 64.62 & 57.93 & 43.86 & 23.68 & 17.87 & 57.11 & 55.34 & 56.21 & 14.29 & 26.76 & 27 \\
    & ST5             & Nucleus 
    & 10.24 & 90.16 & 66.66 & 60.81 & 45.91 & 26.24 & 20.15 & 59.20 & 57.52 & 58.35 & 16.81 & 24.79 & 24 \\
    & MPNet            & 
    & 10.32 & 87.83 & 60.84 & 55.95 & 41.37 & 22.20 & 16.76 & 54.17 & 52.75 & 53.45 & 14.72 & 26.57 & 27 \\
    
     \cmidrule(lr){2-16}
    & SRoBERTa         & 
    & 4.99 & 88.07 & 59.54 & 56.04 & 35.47 & 20.37 & 15.66 & 58.41 & 48.91 & 53.24 & 13.41 & 26.34 & 26  \\
    & SimCSE-BERT     & GPT-2\textsubscript{random}
    & 6.29 & 91.28 & 72.38 & 65.33 & 46.93 & 28.99 & 22.85 & 66.95 & 59.69 & 63.11 & 18.96 & 23.18 & 22  \\
    & SimCSE-RoBERTa  
    & ------------------ 
    & 5.98 & 91.33 & 68.78 & 62.42 & 43.41 & 25.66 & 19.82 & 64.27 & 56.66 & 60.22 & 16.04 & 24.30 & 23 \\
    & ST5             & Beam 
    & 5.90 & 91.47 & 70.72 & 65.45 & 44.52 & 27.83 & 21.99 & 67.46 & 58.26 & 62.53 & 19.11 & 22.68 & 21 \\
    & MPNet            & 
    & 5.64 & 89.27 & 65.08 & 60.39 & 40.04 & 23.83 & 18.54 & 62.64 & 53.51 & 57.72 & 16.63 & 24.66 & 23 \\

    \midrule
    \midrule  
    \multirow{20}{*}{QNLI} 
    & SRoBERTa        & & 25.48 & 80.13 & 26.65 & 22.45 & 20.32 & 3.95 & 2.47 & 28.40 & 27.09 & 27.73 & 0.17 & 95.99 & 75 \\
    & SimCSE-BERT     & GPT-2\textsubscript{pretrain}
    & 26.36 & 80.45 & 30.71 & 26.22 & 23.03 & 5.21 & 3.50 & 30.27 & 29.07 & 29.66 & 0.21 & 91.21 & 70 \\
    & SimCSE-RoBERTa  
    & ------------------
    & 22.85 & 80.13 & 27.85 & 23.76 & 20.54 & 4.26 & 2.75 & 29.12 & 27.26 & 28.16 & 0.22 & 94.97 & 74 \\
    & ST5             & Nucleus
    & 28.74 & 83.21 & 32.77 & 27.86 & 24.11 & 5.86 & 3.91 & 31.84 & 30.69 & 31.25 & 0.37 & 90.71 & 71 \\
    & MPNet           & & 28.66 & 81.46 & 29.94 & 25.11 & 22.57 & 4.88 & 3.19 & 30.38 & 29.23 & 29.80 & 0.33 & 93.72 & 74 \\
    
    \cmidrule(lr){2-16}
    & SRoBERTa        & & 11.09 & 81.18 & 30.54 & 26.23 & 20.40 & 4.97 & 3.28 & 35.02 & 27.33 & 30.70 & 0.32 & 87.47 & 67 \\
    & SimCSE-BERT     & GPT-2\textsubscript{pretrain}
    & 11.74 & 81.37 & 34.24 & 29.83 & 22.63 & 6.24 & 4.35 & 36.84 & 28.87 & 32.37 & 0.50 & 84.77 & 65 \\
    & SimCSE-RoBERTa  
    & ------------------
    & 10.43 & 81.34 & 31.69 & 27.46 & 20.50 & 5.27 & 3.53 & 35.69 & 27.47 & 31.04 & 0.41 & 87.13 & 67 \\
    & ST5             & Beam
    & 12.06 & 84.15 & 36.58 & 31.75 & 23.31 & 6.82 & 4.70 & 39.05 & 30.30 & 34.13 & 0.75 & 83.79 & 64 \\
    & MPNet           & & 12.59 & 82.52 & 33.88 & 28.92 & 22.21 & 5.84 & 3.94 & 37.18 & 29.23 & 32.73 & 0.58 & 85.97 & 66 \\
    
    \cmidrule(lr){2-16}
    & SRoBERTa         & 
    & 91.71 & 80.17 & 31.75 & 26.94 & 23.98 & 5.68 & 3.65 & 32.61 & 30.34 & 31.44 & 0.36 & 89.81 & 67 \\
    & SimCSE-BERT     & GPT-2\textsubscript{random}
    & 94.51 & 81.79 & 37.83 & 32.31 & 27.99 & 7.73 & 5.33 & 36.15 & 33.80 & 34.94 & 0.61 & 86.01 & 63 \\
    & SimCSE-RoBERTa  
    & ------------------ 
    & 78.12 & 82.56 & 36.90 & 31.31 & 26.35 & 7.20 & 4.98 & 37.06 & 33.31 & 35.09 & 0.53 & 86.99 & 64 \\
    & ST5             & Nucleus 
    & 98.87 & 81.43 & 36.05 & 31.14 & 26.73 & 7.31 & 4.90 & 34.84 & 32.60 & 33.68 & 0.41 & 85.86 & 63 \\
    & MPNet            & 
    & 97.55 & 79.81 & 33.55 & 28.52 & 25.06 & 6.17 & 4.11 & 33.24 & 31.08 & 32.12 & 0.46 & 87.85 & 66 \\
    
     \cmidrule(lr){2-16}
    & SRoBERTa         & 
    & 11.13 & 81.04 & 34.34 & 29.97 & 18.52 & 5.48 & 3.67 & 43.81 & 27.19 & 33.56 & 0.50 & 84.79 & 61 \\
    & SimCSE-BERT     & GPT-2\textsubscript{random}
    & 10.83 & 82.24 & 40.01 & 35.25 & 20.58 & 7.19 & 5.08 & 48.78 & 29.49 & 36.76 & 0.85 & 83.14 & 58 \\
    & SimCSE-RoBERTa  
    & ------------------ 
    & 10.69 & 83.12 & 38.73 & 33.76 & 19.47 & 6.54 & 4.63 & 48.62 & 29.26 & 36.53 & 0.81 & 83.88 & 60 \\
    & ST5             & Beam 
    & 10.81 & 82.05 & 38.19 & 33.93 & 19.53 & 6.81 & 4.77 & 47.42 & 28.43 & 35.55 & 0.52 & 83.32 & 59 \\
    & MPNet            & 
    & 11.41 & 80.74 & 36.18 & 31.65 & 19.04 & 5.98 & 4.15 & 44.89 & 27.74 & 34.29 & 0.60 & 84.17 & 60 \\

    \bottomrule
  \end{tabular}
  \vspace{-0.05in}
  \caption{\label{tab:decode methods}
A complete evaluation of generation quality with different decoding algorithms and model initializations.
EMR refers to the exact match ratio and ED stands for the edit distance.
Embedding Similarity, ROUGE, BLEU, Pre, Rec, F1 and EMR are measured in \%.
To make inverted results close to the original inputs, a smaller ED with a larger EMR is expected for embedding inversion attacks.
}
\vspace{-0.05in}
\end{sidewaystable*}

\end{document}